\documentclass[10pt,journal,compsoc,pdftex]{IEEEtran}

\usepackage{hyperref}
\hypersetup{
    colorlinks=true,
    linkcolor=red,
    filecolor=magenta,      
    urlcolor=cyan,
    pdfpagemode=FullScreen,
    }

\usepackage[utf8]{inputenc} 
\usepackage[T1]{fontenc}    
\usepackage{amsfonts}       
\usepackage{nicefrac}       
\usepackage{microtype}      

\usepackage{times}
\usepackage{epsfig}
\usepackage{graphicx}
\usepackage{wrapfig} 
\usepackage{amsmath}
\usepackage{amssymb}

\usepackage{mathtools}
\usepackage{amsthm}

\DeclareMathAlphabet{\mathpzc}{OT1}{pzc}{m}{it}
\usepackage{dsfont} 
\usepackage{xcolor}         
\usepackage[export]{adjustbox} 
\usepackage{multirow}
\usepackage{siunitx}
\usepackage[absolute,overlay]{textpos}
\usepackage{subcaption}
\usepackage{booktabs}

\usepackage{listings}

\definecolor{codekeyword}{rgb}{0.0,0.0,0.6}
\definecolor{codestring}{rgb}{0.58,0.0,0.13}
\definecolor{codecomment}{rgb}{0.25,0.5,0.35}

\lstdefinestyle{pythonstyle}{
    language=Python,
    frame=lines,                 
    framesep=2mm,
    basicstyle=\ttfamily\footnotesize,
    keywordstyle=\color{codekeyword}\bfseries,
    stringstyle=\color{codestring},
    commentstyle=\color{codecomment}\itshape,
    showstringspaces=false,
    breaklines=true,
    captionpos=b,
    columns=fullflexible,
    upquote=true,
}
\usepackage{colortbl}

\usepackage{etoolbox} 

\usepackage{xspace}

\newcommand{\loss}{HOC\xspace}
\newcommand{\dist}{{\rm d}\xspace}

\newtheorem{defn}{Definition}

\newtheorem*{thm*}{Theorem}
\newtheorem*{prop*}{Proposition}
\newtheorem*{lem*}{Lemma}

\newtheorem{innercustomgeneric}{\customgenericname}
\providecommand{\customgenericname}{}
\newcommand{\newcustomtheorem}[2]{%
  \newenvironment{#1}[1]
  {%
   \renewcommand\customgenericname{#2}%
   \renewcommand\theinnercustomgeneric{##1}%
   \innercustomgeneric
  }
  {\endinnercustomgeneric}
}
\newcustomtheorem{customthm}{Theorem}
\newcustomtheorem{customcor}{Corollary}
\newcustomtheorem{customprop}{Proposition}
\newcustomtheorem{customlemma}{Lemma}

\usepackage[capitalize,noabbrev]{cleveref}

\begin{document}

\title{A Stationary (and Therefore Compatible) Representation is All You Need}

\author{Niccolò~Biondi, 
Federico~Pernici,
Simone~Ricci,
and~Alberto~Del~Bimbo,~\IEEEmembership{Senior~Member,~IEEE}
\thanks{
The authors are with Media Integration and Communication Center (MICC), Dipartimento di Ingegneria dell'Informazione, Università degli Studi di Firenze, 50139, Firenze, Italy. E-mail: name.surname@unifi.it
}%
}

\makeatletter
\newcommand{\vst}{\bBigg@{2}}
\newcommand{\vast}{\bBigg@{3.5}}
\newcommand{\Vast}{\bBigg@{5}}
\makeatother

\IEEEtitleabstractindextext{%
\begin{abstract}
Learning compatible representations aims to learn feature representations that can be used interchangeably over time whenever a model undergoes updates. 
In this paper, we demonstrate that stationary representations learned by $d$-Simplex fixed classifiers imply compatibility as in its formal definition. 
This result establishes a foundation for future works and can be directly exploited in practical learning scenarios. We address the challenge of learning compatibility using $d$-Simplex fixed classifiers when the model is sequentially fine-tuned. 
Learning according to a $d$-Simplex fixed classifier with the cross-entropy loss aligns feature distributions at the first-order statistics. Consequently, it may not fully capture higher-order dependencies in the representation between model updates.
To address this issue, we demonstrate that training the model using a $d$-Simplex fixed classifier through a convex combination of the cross-entropy loss and a contrastive loss not only captures higher-order dependencies, but is also equivalent to learning with the cross-entropy under the compatibility constraints. 
We confirm our findings with extensive experiments also considering a new scenario where a pre-trained model is sequentially fine-tuned and occasionally replaced with an improved model. 
We show that stationary representations enable uninterrupted retrieval services (without reprocessing gallery images) while improving performance during model updates and replacements, achieving state-of-the-art. 
Code at: \href{https://github.com/miccunifi/iamcl2r}{GitHub repository}.
\end{abstract}

\begin{IEEEkeywords}
Deep Learning, Representation Learning, Compatible Learning, Lifelong Learning, Fixed Classifiers, Neural Collapse.
\end{IEEEkeywords}
}

\maketitle

\IEEEdisplaynontitleabstractindextext

\IEEEpeerreviewmaketitle

\section{Introduction} \label{sec:intro}

\IEEEPARstart{R}{epresentation} learning is a fundamental aspect of Deep Learning and underpins core tasks such as search, retrieval, and recognition. 
Early research in these areas has largely considered feature representations learned using training data that do not expand over time (see e.g.~\cite{bengio2013representation, geng2020recent, chen2022deep}). 
Less attention has been paid to learning representations where new training data are produced over time, requiring updates to the model (see e.g.,~\cite{pu2021lifelong, ExplFineGrained, zhao2021continual, Wan_2022_CVPR, pu2023memorizing}). 
As this second case has gained increasing research attention, learning compatible representations~\cite{shen2020towards} has emerged as a key issue, addressing the challenging problem of learning a new representation model while avoiding the need to recalculate gallery data features using the updated model.
In fact, re-indexing of the gallery can be computationally prohibitive for large gallery sets and may become infeasible if the original data are no longer available due to privacy concerns or restrictions on data storage ~\cite{sorscher2022beyond, strubell2019energy, price2019privacy}. 

Research on learning compatible representations has considered two distinct approaches for model updating. In the first, addressed by the majority, the new representation is learned by retraining the model from scratch using both new and old training data~\cite{shen2020towards, wang2020unified, biondi2023cores, meng2021learning, zhang2022towards, duggal2021compatibility, pan2023boundary, zhang2021hot, bai2023dual}. In the second, the new representation is learned by sequentially fine-tuning the model with new training data~\cite{iscen2020memory, Wan_2022_CVPR, biondi2022cl2r, bui2025learning, chen2024anti, goswami2025query}.  
Recently, research in~\cite{biondi2023cores} and~\cite{biondi2022cl2r} has provided empirical evidence that training using the $d$-Simplex fixed classifier achieves state-of-the-art performance in learning compatible representation in both approaches. 
In the $d$-Simplex fixed classifier, class prototypes are set at the vertices of the regular simplex polytope. This results into maximally-separated equiangular prototypes and stationary features~\cite{pernici2021regular}.

However, despite substantial empirical support for learning compatible representations, further research is needed to bridge the gap between empirical outcomes and theoretical analysis.
In this paper, we demonstrate that the representation as learned according to the $d$-Simplex fixed classifier satisfies the compatibility inequalities as defined in~\cite{shen2020towards}.
The demonstration relies on the assumption of the Unconstrained Feature Model (UFM)~\cite{mixon2022neural,zhu2021geometric} and exploits the fact that, using the $d$-Simplex, the learned representation remains stationary during the learning process. 

This theoretical result not only establishes a solid foundation for future research but also has practical applications in compatible representation learning. 
In particular, we address the scenario in compatible representation learning where the new representation is learned by sequentially fine-tuning the model with new data.
In this scenario, a specific challenge is the tendency of both the old and the newly fine-tuned models to align based on their first-order statistics, a characteristic of stationary representations. As a result, relying solely on cross-entropy-based prediction errors during fine-tuning might not adequately capture higher-order dependencies. To address this issue and preserve compatibility, we train the model with a $d$-Simplex fixed classifier using a convex combination of cross-entropy loss and contrastive loss. 
We demonstrate this approach also captures higher-order dependencies in the learned representation and is equivalent to learning under the compatibility inequality constraints as defined in~\cite{shen2020towards}.  

We validate our findings through comprehensive experiments in open-set image recognition, achieving state-of-the-art performance in learning compatible representations. We also consider a novel scenario in which a pre-trained model is sequentially fine-tuned and occasionally replaced by an improved model. 

This paper extends the preliminary results from \cite{biondi2024stationary}, which explored the optimal compatibility approximation using stationary representations. In that earlier study, only one of the two compatibility inequalities was verified. 
Here, we prove that stationarity implies compatibility \textit{without} approximation, verifying both the inequalities of the definition. This is accomplished by explicitly leveraging the cosine distance in the hyperspherical representation space. 
The paper further extends the experimental evaluation also in the compatible lifelong learning representations scenario with other state-of-the-art methods.

This paper is structured as follows.
Sec.~\ref{sec:related_works} provides an overview of relevant literature.
Next, in Sec.~\ref{sec:proof}, we demonstrate that stationary representations learned through the $d$-Simplex fixed classifiers imply compatibility.
In Sec.~\ref{sec:method}, we demonstrate that training the model using a $d$-Simplex with a convex combination of cross-entropy and contrastive loss is equivalent to learning under compatibility constraints.
In Sec.~\ref{sec:experimentalResults}, we present the two evaluation scenarios and comparative evaluations. 
In Sec.~\ref{sec:ablation}, we report ablation studies. 
Conclusions are provided in Sec.~\ref{sec:conclusion_future_directions}.
Proofs and implementation details are provided in the appendices.

\section{Related Work}\label{sec:related_works}

\noindent\textbf{Compatible Representation Learning. \;}
First research in compatible representation learning investigated scenarios where the model is retrained from scratch with both old and new data to learn new representations.
Pioneering research in~\cite{shen2020towards} proposed Backward Compatible Training (BCT). In this approach, compatibility is achieved by directly comparing new and old representations, using the old classifier as a regularizer.
This approach was extended in several works~\cite{zhang2022towards, duggal2021compatibility, pan2023boundary}. 
In UniBCT~\cite{zhang2022towards}, compatibility was achieved by refining the class prototype neighborhood with a fully-connected graph. 
In~\cite{duggal2021compatibility}, it was considered the case in which gallery features are generated from a large model and query features are extracted using a smaller model. 
Recently, the authors in~\cite{pan2023boundary} proposed AdvBCT that uses adversarial learning to minimize the distribution disparity between old and new representations and an elastic boundary constraint to improve compatibility and discrimination.
In~\cite{biondi2023cores}, the authors presented CoReS that exploits feature stationarity provided by the $d$-Simplex fixed classifier to obtain model compatibility.
Other approaches train a lightweight transformation to convert old representations into new ones for compatibility~\cite{Chen_2019_CVPR, wang2020unified, ramanujan2021forward,meng2021learning, ricci2025orthogonality}. 
The authors of LCE~\cite{meng2021learning} achieved compatibility by aligning the class prototypes across models by learning transformations or through a direct comparison using an identity mapping.

More recently, there have been a few works addressing compatible representation learning where the new representation is learned by sequentially fine-tuning the model with new data.
In FAN~\cite{iscen2020memory}, the authors followed the mapping-based approach and learned a feature-adaptation network to map old features in the representation space of the updated model.
In CVS~\cite{Wan_2022_CVPR}, a composition of distillation functions was applied to the network output to align the updated model with the old model while being coherent with all previous representations.
In~\cite{biondi2022cl2r}, the authors extended their previous research in~\cite{biondi2023cores} by formalizing the Compatible Lifelong Learning Representations (CL$^2$R) scenario. They also showed that using $d$-Simplex fixed classifiers with mean square error loss maintains compatibility when updating the model. 
In \cite{bui2025learning}, features are mapped in a hyperbolic space and used a contrastive loss to constrain the updated features to lie within the entailment cone of the old ones.

While all these papers present empirical evidence of the effectiveness of their solutions, the work in \cite{biondi2024stationary} presents a theoretical demonstration. 
It shows that learning with $d$-Simplex fixed classifiers optimally approximates the compatibility inequalities of~\cite{shen2020towards}. 
It was proved that only the first compatibility inequality is satisfied, whereas the second is not. Despite of that, prototypes within the $d$-Simplex fixed classifier arrange at their maximum pairwise distance, which is the optimal relative distance achievable. 
Here, we prove that stationarity as provided by $d$-Simplex fixed classifiers satisfies both compatibility inequalities \textit{without} approximation. 
This is due to a different geometric assumption. 
In \cite{biondi2024stationary}, representations were modeled as hyperballs and the Euclidean distance was used as the distance metric. Here, we model representations as hyperspherical caps lying on a unit hypersphere and cosine distance is used as distance metric.
\\

\noindent\textbf{Neural Collapse. \;} 
Neural Collapse~\cite{papyan2020prevalence} is a phenomenon that occurs in the terminal phase of training of any deep neural network, where feature vectors of the same class collapse to their class prototypes~\cite{kothapalli2022neural}. 
This leads to a configuration where class prototypes form a Simplex Equiangular Tight Frame (Simplex-ETF), i.e., such that their pairwise angles are the same~\cite{fickus2018equiangular}.
A simplified model, introduced in~\cite{mixon2022neural} and \cite{fang2021exploring} under the name of Unconstrained Feature Model (UFM) and Layered Peeled Model (LPM), respectively, was proposed to analyze the geometries of the feature space and the training dynamics of Neural Collapse. 
In this model, assuming that the neural network has enough expressivity, the penultimate layer of the network can be regarded as disconnected from the previous layers and features can be treated as free optimization variables during training. Using UFM, several authors have shown that Neural Collapse also occurs with different types of loss and constraints~\cite{ji2021unconstrained, zhu2021geometric, yang2022we, han2021neural}. 
In particular, in~\cite{zhu2021geometric,yang2022we}, it was demonstrated that Neural Collapse emerges when a classifier is initially configured as a Simplex-ETF and kept fixed during training, also when the dataset has class imbalance. 

Before Neural Collapse was observed, earlier studies already configured classifiers with the Simplex-ETF fixed geometry at the beginning of training \cite{Pernici_2019_CVPR_Workshops, pernici2021regular}. Fixing the geometry beforehand enables the preservation of future class regions in representation space, as introduced in \cite{pernici2021class} and further explored in \cite{yang2022neural} and \cite{zhou2022forward}.
Our proof relies on the UFM and LPM assumption that the backbone is expressive enough for independent feature analysis and the $d$-Simplex ETF fixed classifier's symmetry. These assumptions reduce the analysis to one class interaction, as it makes all interactions identical.

\section{Stationarity Implies Compatibility} 
\label{sec:proof}

Let \(\mathcal{G} = \{(\mathbf{x}_i,y_i)\}_{i=1}^{N_g}\)  be a gallery set of \(N_g\) images \(\mathbf{x}_i \in \mathbb{R}^D\), each with its corresponding class label \(y_i \in \mathcal{Y}\), and \(\Phi^\mathcal{G} = \{\phi(\mathbf{x}_i) \in \mathbb{R}^d \, | \, \forall \mathbf{x}_i \in \mathcal{G}\}\) the set of feature vectors for the gallery set, obtained with the representation model
$\phi$. Similarly, we denote \(\mathcal{Q} = \{\mathbf{x}_i\}_{i=1}^{N_q}\) a query set composed of $N_q$ images $\mathbf{x}_i$ and \(\Phi^\mathcal{Q} = \{\phi(\mathbf{x}_i) \in \mathbb{R}^d \, | \, \forall \mathbf{x}_i \in \mathcal{Q}\}\) the set of query features vectors obtained with $\phi$.

We consider a sequence of \(T\) training task-sets \( (\mathcal{T}^1, \mathcal{T}^2, \ldots, \mathcal{T}^T) \), where \(\mathcal{T}^t = \{ (\mathbf{x}_i, y_i) \}_{i=1}^{N_t}\) includes \(N_t\) training instances $\mathbf{x}_i$ each with its class label \(y_i \in \mathcal{Y}_t\).
At each task $t$, a new model $\phi_t$ is learned either by fine-tuning the previous representation model $\phi_{t-1}$ using the task-set $\mathcal{T}^t$ trying to preserve the knowledge from the previous tasks or by training from scratch using all the task-sets up to the $t$-th task. 

Search/retrieval is performed by finding the nearest gallery features to a given query feature, measured by a distance function $\dist(\cdot, \cdot)$. A new model $\phi_t$ is compatible with a previous model $\phi_k$ if its query features in $\Phi_t^\mathcal{Q}$ can be directly compared to gallery features in $\Phi_k^\mathcal{G}$. This avoids costly re-indexing of the gallery feature set after every model update.
More formally, compatibility is defined as follows~\cite{shen2020towards}:
\begin{defn}[Compatibility] \label{def:compatibility-shen}
Given two representation models \(\phi_k\) and \(\phi_t\), where \(\phi_t\) is learned after \(\phi_k\), the models are deemed compatible with respect to a distance function \(\dist(\cdot, \cdot)\) if the following conditions are satisfied for all pairs of samples \(\mathbf{x}_i\) and \(\mathbf{x}_j\):
\begin{subequations}\label{eq:compatible_set_dist}
\begin{align}
 &\dist \big(\phi_{k}(\mathbf{x}_i), \phi_{t}(\mathbf{x}_j) \big) \leq \dist \big(\phi_{k}(\mathbf{x}_i), \phi_{k}(\mathbf{x}_j) \big), & \text{\rm for } y_i = y_j, \label{eq:first} \\
 &\dist \big(\phi_{k}(\mathbf{x}_i), \phi_{t}(\mathbf{x}_j) \big) \geq \dist \big(\phi_{k}(\mathbf{x}_i), \phi_{k}(\mathbf{x}_j) \big), & \text{\rm for } y_i \neq y_j, \label{eq:second}
\end{align}
\end{subequations}
\( \text{\rm for } t \in \{2,3, \dots, T\} \,  \text{\rm and }  k \in \{1,2, \dots, T-1\}\),   \( \text{\rm with } k < t\).
\end{defn}

\begin{figure}
    \centering
    \begin{subfigure}{0.39\linewidth}
        \centering
        \includegraphics[width=\linewidth]{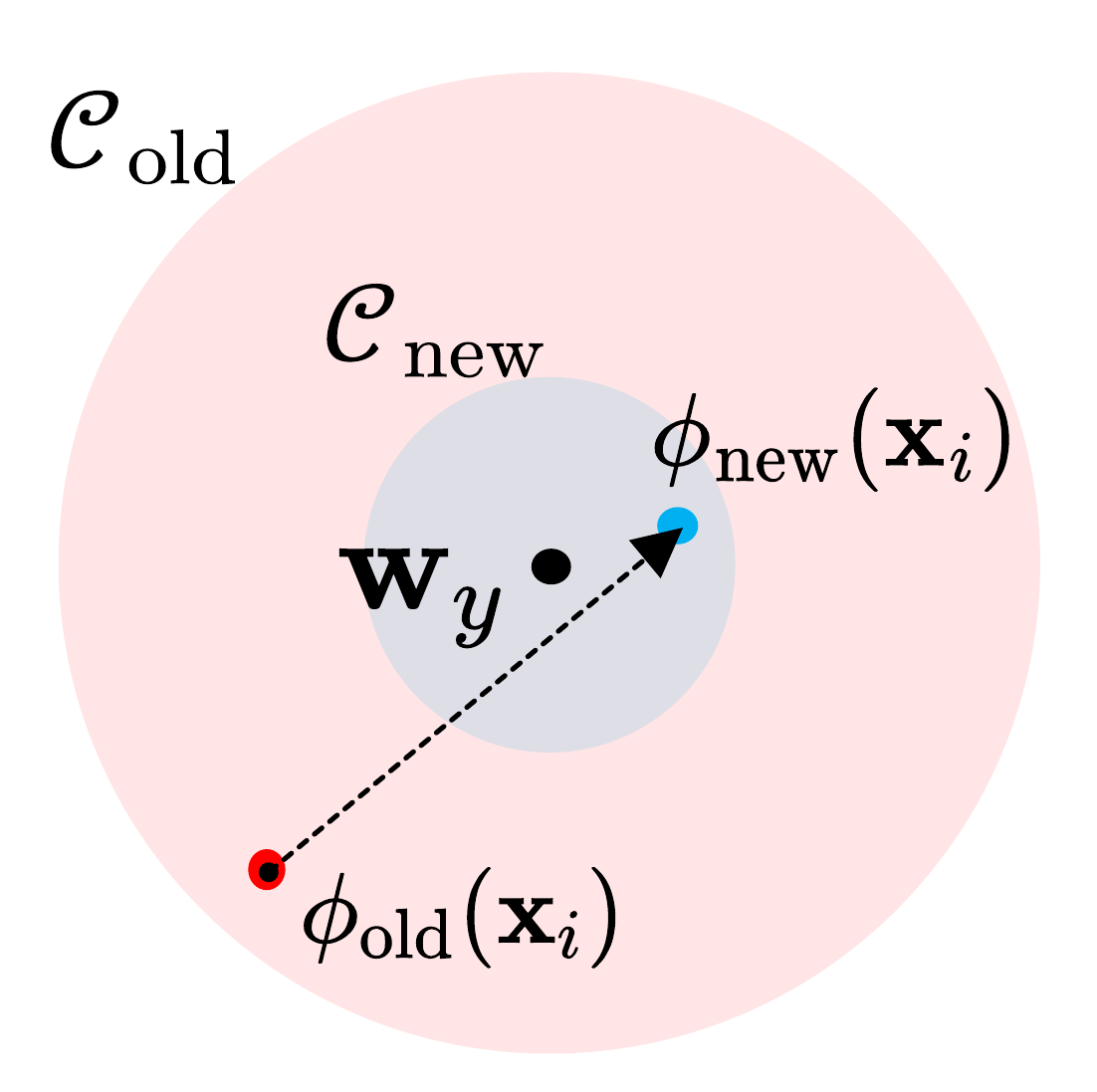}
        \caption{Model update for a feature sample.}
        \label{fig:sa}
    \end{subfigure}
    \hspace{15pt}
    \begin{subfigure}{0.39\linewidth}
        \centering
        \includegraphics[width=0.99\linewidth]{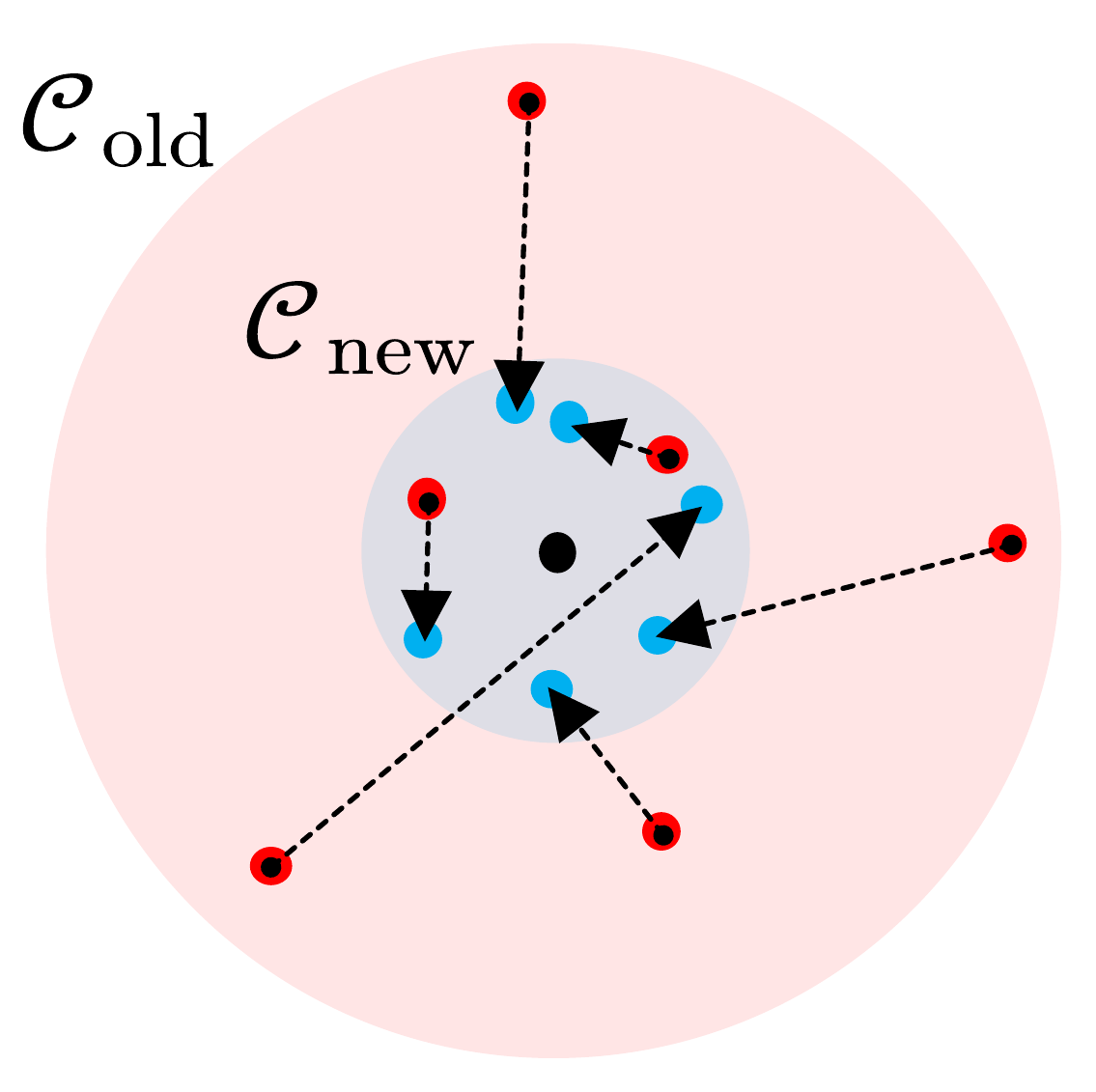}
        \caption{Model update for multiple feature samples.}
        \label{fig:sb}
    \end{subfigure}
    \caption{Key concepts used in Theorem~\ref{theo:compatibility} (illustrated in 2D). Concentric circles represent hyperspherical caps for a generic class $y$ as learned before and after the update  according to a $d$-Simplex fixed classifier. Due the stationarity provided by the $d$-Simplex fixed classifier circles are concentric (in red and cyan before and after the update, respectively)}.
    \label{fig:proof}
\end{figure}
\begin{figure*}[t]
    \centering    
    \hspace{-25pt}
    \begin{subfigure}{0.52\linewidth}
    \centering
    \adjincludegraphics[height=4.2cm,trim={0 0 {.65\width} 0},clip]   {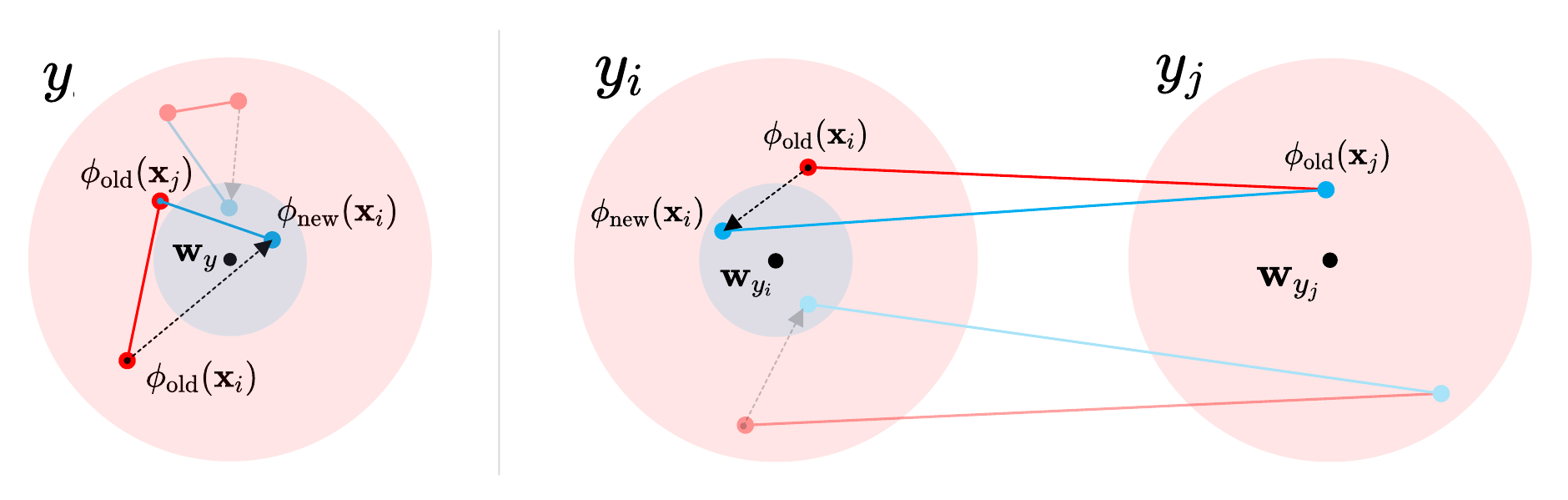}
    \caption{Same class}\label{fig:feat_space_change_pre}
    \end{subfigure}
    \hspace{-98pt}
    \begin{subfigure}{0.55\linewidth}
    \centering        
    \adjincludegraphics[height=4.2cm,trim={{.35\width} 0 0 0},clip]        {images/proof-fig.pdf}
    \caption{Different classes}
    \label{fig:feat_space_change_post}
    \end{subfigure}
    \caption{Conditions under which Theorem~\ref{theo:compatibility} is demonstrated (illustrated in 2D).
    (a): Two distinct instances $\mathbf{x}_i$ and $\mathbf{x}_j$ of the same class and their distances before (red) and after (cyan) the update (Eq.~\ref{eq:first}).
    (b): Two distinct instances $\mathbf{x}_i$ and $\mathbf{x}_j$ of different classes and their distances before (red) and after (cyan) the update (Eq.~\ref{eq:second}).  Compatibility is verified by computing the expected lengths of the segments and verifying if they satisfy the inequalities of Def. \ref{def:compatibility-shen}. Despite inequalities are not satisfied for some instances (in transparent colors), they are verified in expectation according to Theorem~\ref{theo:compatibility}.
    }
    \label{fig:theo-diff-classes-dist}
\end{figure*}

Let us now consider the case in which the model is trained according to a $d$-Simplex fixed classifier and a large number of classes is pre-allocated to reserve regions in the feature space for future classes. In a feature space of dimension $d$, the number of $d$-Simplex classes $K$ equals the number of the $d$-Simplex vertices, i.e., $K = d+1$~\cite{pernici2021regular}. 
Using this configuration, class prototypes are equidistant from each other and maximally separated and learned features are stationary~\cite{Pernici_2019_CVPR_Workshops}. 

The following Theorem~\ref{theo:compatibility} provides a theoretical foundation for compatibility by relating feature stationarity, as learned by the $d$-Simplex fixed classifier, to compatibility, as defined in Def.~\ref{def:compatibility-shen}.

\begin{customthm}{1}[Stationarity implies Compatibility] 
\label{theo:compatibility}
Let $\phi_k$ and $\phi_t$ be representation models, respectively learned at the $k$-th and $t$-th tasks, by a $d$-Simplex fixed classifier with $K$ pre-allocated classes and class prototypes $\mathbf{w}_1, \mathbf{w}_2, \ldots, \mathbf{w}_K$. The classes learned by $\phi_k$ and $\phi_t$ are respectively $K_k$ and $K_t$, with $K_k < K_t < K$.
Under the assumption that class hyperspherical caps shrink after model updates, it follows that $\phi_{k}$  and $\phi_{t}$ satisfy on average the compatibility inequalities as in Def.~\ref{def:compatibility-shen}.
\end{customthm}

\noindent The proof of Theorem~\ref{theo:compatibility} is provided in Appendix~\ref{sec:proof-app}. 

Fig.~\ref{fig:proof} and Fig.~\ref{fig:theo-diff-classes-dist} illustrate key concepts and relationships used in Theorem~\ref{theo:compatibility}. Without loss of generality, we consider two distinct hyperspherical caps $\mathcal{C}_{\rm old}(\mathbf{w}_y, \theta_{\rm old}^y)$ and $\mathcal{C}_{\rm new}(\mathbf{w}_y, \theta_{\rm new}^y)$ of a generic class $y$, before and after a learning update, respectively. 
For ease of reading, in Fig.~\ref{fig:proof} these caps are represented as simple 2D discs, and their angles ($\theta_{\rm new}^y$ and $\theta_{\rm old}^y$, respectively) are related to their radii.
Due to the stationarity provided by $d$-Simplex classifier, the hyperspherical caps $\mathcal{C}_{\rm new}(\mathbf{w}_y, \theta_{\rm new}^y)$ and $\mathcal{C}_{\rm old}(\mathbf{w}_y, \theta_{\rm old}^y)$ align along the same central axis in the representation space.
After the learning step, $\mathcal{C}_{\rm new}(\mathbf{w}_y, \theta_{\rm new}^y)$ has a reduced angle (adding new information improves the discrimination capability of the model~\cite{prato2021scaling,caballero2023broken}) than $\mathcal{C}_{\rm old}(\mathbf{w}_y, \theta_{\rm old}^y)$. 

Fig.~\ref{fig:theo-diff-classes-dist} instead provides a graphical sketch of conditions under which compatibility is demonstrated in Theorem \ref{theo:compatibility}.
The figure shows in lighter colors sample instances where Eq. \ref{eq:first} and Eq. \ref{eq:second} do not hold.
Theorem~\ref{theo:compatibility} demonstrates that although some instances may not comply with the compatibility constraints, nevertheless, under the specified assumptions, these constraints are satisfied in expectation.

\section{Stationarity and Compatibility in Sequential Fine-tuning} 
\label{sec:method}
We consider in the following the case in which the new representation is learned by sequential fine-tuning. 
In this case, at each task, the average prediction errors of the representation model $\phi_t$ are computed with respect to a previously learned model $\phi_{t-1}$.
As shown in~\cite{pernici2021regular}, training a model through a $d$-Simplex fixed classifier with cross-entropy loss aligns features of each class to the first-order statistics of their distribution that are kept in the same position across model updates.
Consequently, first-order statistics before and after the updates are aligned. According to this, the average prediction errors may fail to capture higher-order dependencies in the representation between model updates, and may result in low values of the cross-entropy loss, thereby limiting backpropagation in properly learning the model.

To address this problem and preserve compatibility, we train the representation model $\phi_t$ using $d$-Simplex with a convex combination of the cross-entropy loss and a contrastive loss, namely: 
\begin{equation} 
\label{eq:total_loss}
\resizebox{0.91\hsize}{!}{$%
    \mathcal{L}_\textsc{hoc} (\phi_t) = \lambda \, \mathcal{L}_{\textsc{sce}}(\phi_t) + (1 - \lambda) \, \mathcal{L}_{i\textsc{nce}} (\phi_t, \phi_{t-1}) \;\; \lambda \in [0,1].
$}
\end{equation}

The term
\begin{equation} \label{eq:loss_ce_simplex} 
\resizebox{0.9\hsize}{!}{$%
\begin{aligned}
    &\mathcal{L}_{\textsc{sce}} (\phi_t)= 
    &= - \sum\limits_{B} \log \!  \vast( 
    \dfrac {\text{exp} \Big( \mathbf{W}_{y_i}^{\top}{{\phi_t(\mathbf{x}_i)}} \Big) } {\sum\limits_{\scriptscriptstyle j =1}^{K_t} \text{exp} \Big( \mathbf{W}_{j}^{\top}{{\phi_t(\mathbf{x}_i)}} \Big)  + \sum\limits_{\scriptscriptstyle j=K_t+1}^{K} \text{exp} \Big( \mathbf{W}_{j}^{\top}{{\phi_t(\mathbf{x}_i)}} \Big) } \vast)
\end{aligned}
$}
\end{equation}
is the $d$-Simplex cross-entropy loss where the first term in the denominator accounts for the classes learned up to the $t$-th task and the second term accounts for the future classes, preserving dedicated regions within the representation space. $\mathbf{W}^{\top}_{j} \in \mathbb{R}^d$ denotes the $j$-th column of the $d$-Simplex classifier matrix $\mathbf{W} \in \mathbb{R}^{d \times K}$, $K_t$ the number of classes learned up to the $t$-th task with $K_t < K$, and $B$ is a mini-batch of samples $\mathbf{x}_i$ drawn from the task-set $\mathcal{T}^{t}$ or, depending on the implementation, it can include samples from an Experience Replay buffer. 

The term
\begin{equation}\label{eq:CONTRAST}
\resizebox{0.91\hsize}{!}{$
    \mathcal{L}_{i\textsc{nce}} (\phi_t, \phi_{t-1}) = - \sum\limits_{B} 
        \log \!  \left( 
        \dfrac {\Delta \big( \phi_{t-1}(\mathbf{x}_i), \phi_t(\mathbf{x}_i) \big)  } { \sum\limits_{j \neq i}  \Delta \big( \phi_{t-1}(\mathbf{x}_i), \phi_t(\mathbf{x}_j) \big)   } \right) 
    $}
\end{equation}
is the contrastive loss~\cite{oord2018representation, tian2019contrastive} defined with respect to the $\rho$-scaled cosine similarity between $\phi_{t-1}(\mathbf{x}_i)$ and $\phi_{t}(\mathbf{x}_j)$
\begin{equation}
\label{eq:CONTRAST2}
\resizebox{0.91\hsize}{!}{$
    \Delta \big( \phi_{t-1}(\mathbf{x}_i), \phi_t(\mathbf{x}_j) \big) = \text{exp} \left( \rho \;\; \frac{\phi_{t-1}(\mathbf{x}_i) \phi_{t}(\mathbf{x}_j)}{ ||\phi_{t-1}(\mathbf{x}_i)|| \, ||\phi_{t}(\mathbf{x}_j)|| } \right).
$}
\end{equation}
It represents an approximation of the Kullback-Leibler divergence between the joint distribution and the product of the marginals of \(\phi_t\) and \(\phi_{t-1}\)~\cite{tian2020contrastive}, therefore also approximating the mutual information between \(\phi_t\) and \(\phi_{t-1}\)~\cite{oord2018representation}. 
Consequently, $\mathcal{L}_{i\textsc{nce}}$ captures the higher-order dependencies in the representation between model updates~\cite{cover1999elements}.

\begin{figure}[t]
    \centering
    \includegraphics[width=0.7\linewidth]{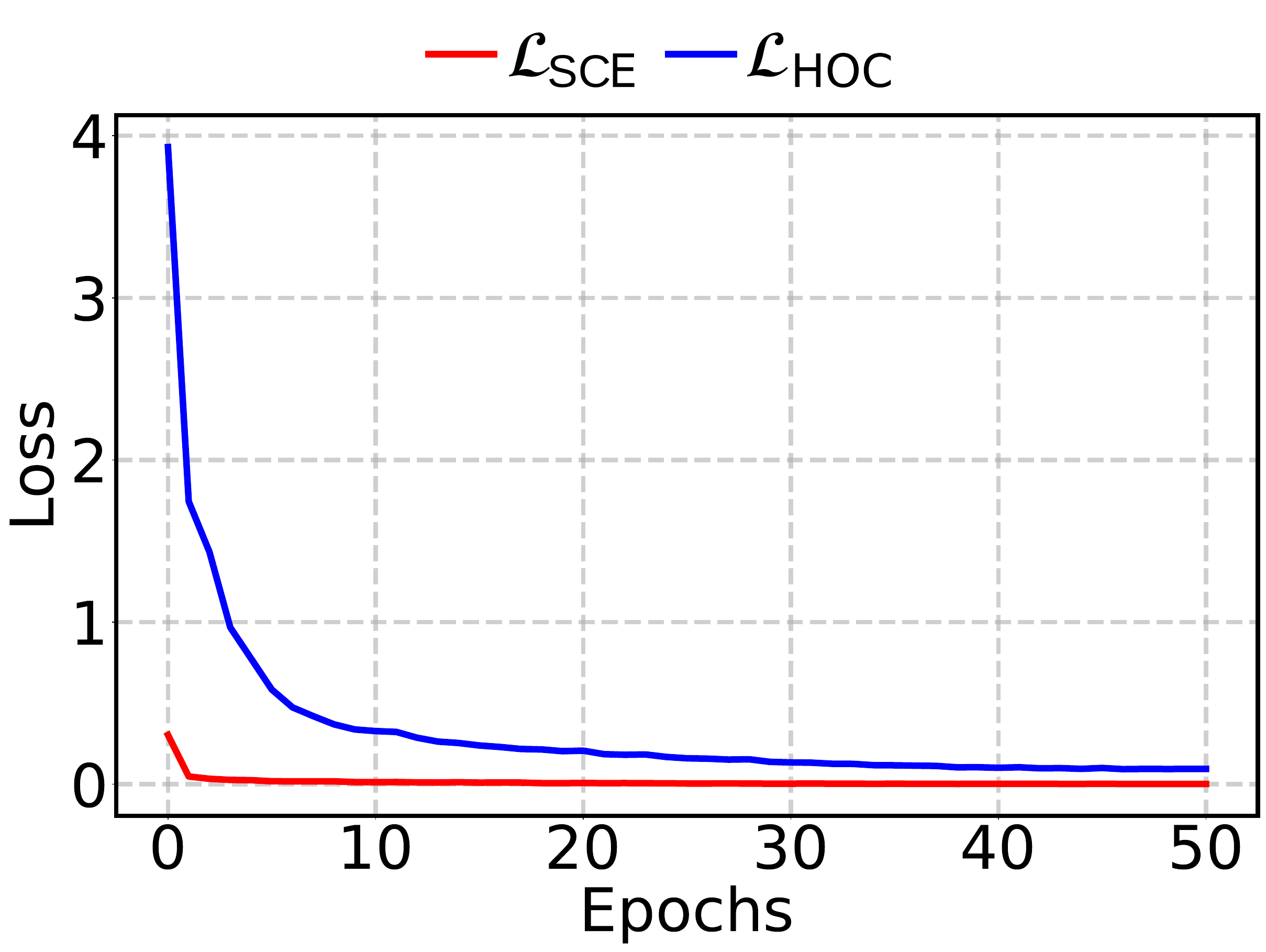}
    \caption{Fine-tuning LeNet++ using $d$-Simplex fixed classifier with two distinct losses: $d$-Simplex cross-entropy loss of Eq.~\ref{eq:loss_ce_simplex} (red line) and HOC loss of Eq.~\ref{eq:total_loss} (blue line). Models are trained on MNIST. 
    }
    \label{fig:toyproblem}
\end{figure}

In the following, we refer to $\mathcal{L}_\textsc{hoc}$ loss as Higher-Order Compatibility (HOC) loss and the representation model trained according to the $d$-Simplex with the $\mathcal{L}_\textsc{hoc}$~loss as $d$-Simplex-HOC.
Fig.~\ref{fig:toyproblem} shows the effect of $\mathcal{L}_\text{\loss}$ to capture higher-order dependencies in comparison with the cross-entropy loss with a toy example. We trained a LeNet++ CNN \cite{discriminative2016wen} with the $d$-Simplex fixed classifier on five MNIST classes and then fine-tuned the model with the other five classes (40 epochs). 
It can be noticed that the training error from cross-entropy (red curve) quickly drops to low values, whereas the convergence using the $\mathcal{L}_\textsc{hoc}$ (blue curve) occurs more slowly, allowing for the capture of more information during back-propagation.

The following proposition demonstrates that, using the $\mathcal{L}_\textsc{hoc}$ loss of Eq.~\ref{eq:total_loss}, while accounting for higher-order dependencies between model updates, still preserves compatibility between learned representations. 

\begin{customprop}{1}\label{theo:equivalence}
Let $\phi_{t}$, $\phi_{t-1}$ be two representation models learned according to a $d$-Simplex fixed classifier where the updated model $\phi_t$ is obtained by fine-tuning $\phi_{t-1}$. Then, training $\phi_t$ with the HOC loss of Eq.~\ref{eq:total_loss} is equivalent to training $\phi_t$ using cross-entropy loss under compatibility constraints.
\end{customprop}

The demonstration of Proposition~\ref{theo:equivalence} is provided in Appendix~\ref{sec:appenix_proof_hoc}. 

\section{Experimental Results} \label{sec:experimentalResults}

In this section, we present empirical evidence and practical implications of the theoretical results presented in Sec.~\ref{sec:proof} and Sec.~\ref{sec:method}. 
We consider two distinct learning scenarios with sequential fine-tuning in visual search applications. 
In the first, we assess compatibility considering a model which is initially trained from scratch and then subjected to sequential fine-tuning~\cite{Wan_2022_CVPR}.
As in \cite{biondi2022cl2r}, we refer to this scenario as \textit{Compatible Lifelong Learning Representation} (CL$^2$R).
In the second, we analyze compatibility in a novel case where a pre-trained model undergoes sequential fine-tuning, but at some times is replaced with an improved model. 
We refer to this scenario as \textit{Improved Asynchronous Model Compatible Lifelong Learning Representation} (IAM-CL$^2$R, read ``I am clear''). 
The experiments follow the 1:N search testing protocol in open-set image recognition, such that the training and test sets do not share any classes.
We perform a comparative analysis of $d$-Simplex-HOC against: a baseline method, named Experience Replay (ER), where the model undergoes fine-tuning using data from the new task and an Experience Replay buffer; FAN~\cite{iscen2020memory}; CVS~\cite{Wan_2022_CVPR}; $d$-Simplex-FD~\cite{biondi2022cl2r};  adapted versions of BCT~\cite{shen2020towards} (BCT-ER), LCE~\cite{meng2021learning} (LCE-ER), and AdvBCT~\cite{pan2023boundary} (AdvBCT-ER), all of which utilize an experience replay buffer to perform sequential fine-tuning.

\subsection{Evaluation Metrics}
To evaluate the compatibility between a model $\phi_t$ and a previously learned model $\phi_k$, we use the Empirical Compatibility Criterion of~\cite{shen2020towards}
\begin{eqnarray} \label{eq:multistepecc}
M \big( \Phi_t^\mathcal{Q}, \Phi_k^\mathcal{G} \big) > 
M \big( \Phi_k^\mathcal{Q}, \Phi_k^\mathcal{G} \big) {\rm \quad with \:} t > k,
\end{eqnarray}
where $M \big( \Phi_t^\mathcal{Q}, \Phi_k^\mathcal{G} \big)$ is the cross-test between $\phi_t$ and $\phi_k$ and $M \big( \Phi_k^\mathcal{Q}, \Phi_k^\mathcal{G} \big)$ the self-test of $\phi_k$, being $\Phi_t^\mathcal{Q}$ and $\Phi_k^\mathcal{Q}$ the set of query features extracted with $\phi_t$ and $\phi_k$, respectively, $\Phi_k^\mathcal{G}$ the set of gallery features extracted with $\phi_k$, and $M$ a performance metric. 
In our experiments, we use the 1:N search (identification) accuracy as the metric $M$ according to the cosine similarity between query and gallery features. 
To represent compatibility across $T$ tasks, we use the Compatibility Matrix $C \in \mathbb{R}^{T \times T}$ as defined in~\cite{biondi2022cl2r}
\begin{equation}
    C_{t, k} =
    \begin{cases}
      \qquad 0 & \text{ if } t < k \\
      M \big( \Phi_k^\mathcal{Q}, \Phi_k^\mathcal{G} \big) & \text{ if } t = k \\
      M \big( \Phi_t^\mathcal{Q}, \Phi_k^\mathcal{G} \big) & \text{ if } t > k 
    \end{cases},
    \label{eq:compatibility_matrix}
\end{equation}   
Each element $C_{t,k}$ of the Compatibility Matrix indicates the performance that is achieved when using model $\phi_t$ for the query features and model $\phi_k$ for the gallery features. Self-tests are reported in the main diagonal, and cross-tests in the lower sub-diagonal matrix.

The following scalar metrics derived from the Compatibility Matrix are used to evaluate compatibility and accuracy under compatible learning
\begin{itemize}   
 \item     \textit{Average Compatibility}~\cite{biondi2023cores}
\begin{equation*}\label{eq:norm_mecc}
            AC = \frac{2}{T(T-1)} \sum\limits_{t=2}^{T} \sum\limits_{k=1}^{t-1} \mathds{1}{ \Big( C_{t,k} >  C_{k,k}\Big) },
    \end{equation*}
    expressing the normalized count of the number of times in which compatibility is assessed over $T$ tasks according to Eq.~\ref{eq:multistepecc}. It serves as a measure of the degree to which compatibility is achieved, taking into account all possible pairs of model combinations up to $T$. 
    \item     \textit{Average Compatibility up to Task $\tau$} \cite{biondi2024stationary}
\begin{equation*}
            AC_{\tau} = \frac{2}{\tau(\tau-1)} \sum\limits_{t=2}^{\tau} \sum\limits_{k=1}^{t-1} \mathds{1}{ \Big( C_{t,k} >  C_{k,k}\Big) },
    \end{equation*}
    expressing the normalized count of the number of times in which compatibility is assessed according to Eq.~\ref{eq:multistepecc} over $\tau$ tasks, with $\tau < T$. 
    \\
\item \textit{Average Accuracy}~\cite{biondi2023cores}
    \begin{align*}
        AA =& \frac{2}{T(T+1)} \sum\limits_{t=1}^{T} \sum_{k=1}^{t} C_{t,k},
    \end{align*}    
    providing the average measure of accuracy over $T$ tasks, obtained from all the possible combinations of previously learned models in self- and cross-tests. 
\item \textit{Average Accuracy up to Task $\tau$} \cite{biondi2024stationary}
    \begin{align*}
        AA_{\tau} =& \frac{2}{\tau(\tau+1)} \sum\limits_{t=1}^{\tau} \sum_{k=1}^{t} C_{t,k}, 
    \end{align*}      
    providing the average value of the accuracy over $\tau$ tasks, obtained from all the possible combinations of previously learned models in self- and cross-tests up to task $\tau < T$.  
\item 
\textit{Average Compatibility Accuracy}

    \begin{align*}
        \mathit{ACA} =& \frac{2}{T(T-1)} \sum\limits_{t=2}^{T} \sum\limits_{k=1}^{t-1} 
        \mathds{1}{ \Big( C_{t,k} >  C_{k,k}\Big) } {,} \\
    \end{align*}  
    providing an average measure of accuracy over $T$ tasks, exclusively for the elements $C_{t,k}$ that satisfy Eq.~\ref{eq:multistepecc}. 
    It allows to evaluate the trade-off between compatibility and accuracy.
\end{itemize}

\begin{table}[t]
\centering
\caption{Settings for the CL$^2$R scenario.}
\label{tab:cl2r_settings}
\setlength{\tabcolsep}{2pt}
\begin{tabular}{@{}lccc@{}}
\toprule
\textsc{Settings} & CIFAR100/10 & TinyImageNet200/20 & CUB180/20 \\
\midrule
\multicolumn{4}{l}{\textit{Datasets}} \\
Training Dataset* & CIFAR100 \cite{Krizhevsky2009LearningML} & TinyImageNet200 \cite{le2015tiny} & CUB180 \\
Test Dataset** & CIFAR10 \cite{Krizhevsky2009LearningML} & TinyImageNet20 & CUB20 \\
\quad - Query Samples & 50,000 & 25,478 & 588\\
\quad - Gallery Samples & 10,000  & 1,000 & 598 \\
\midrule
\multicolumn{4}{l}{\textit{Network Architecture}} \\
ResNet Backbone \cite{resnet} & RN32 & RN18 & RN50 \\
Input Size & 32$\times$32 & 64$\times$64 & 224$\times$224 \\
\midrule
\multicolumn{4}{l}{\textit{Feature Dimensions}} \\
$d$-Simplex ($K-1$) & 99 ($K$=100)   & 199  ($K$=200)  & 179  ($K$=180) \\
Compared Methods*** & 64 & 512 & 2048 \\
\midrule
\multicolumn{4}{l}{\textit{Training Hyperparameters}} \\
Optimizer & SGD & SGD & SGD \\
Epochs & 70 & 90 & 150 \\
Batch Size & 128 & 256 & 64 \\
Initial LR & 0.1 & 0.1 & 0.01 \\
LR Decay Epochs & 50, 64 & 30, 60 & 60, 90, 120 \\
\bottomrule
\end{tabular}%
\begin{flushleft}
* The original CUB200 dataset \cite{wah2011caltech} has been split into CUB180 for training and CUB20 for test. \\
** The test set contains classes disjoint from the training set. The training and test splits of the test datasets are used for query and gallery, respectively. The list of classes used for the TinyImageNet20 and CUB20 splits is provided in Appendix~\ref{sec:indices_classes}.\\
*** The feature size corresponds to the size of the penultimate layer of the backbone.
\end{flushleft}
\end{table}

\begin{table*}[t]
 \caption{1:N search CL$^2$R scenario. Performance metrics with CIFAR100/10 for 2, 7, 16, and 31 tasks. Number of classes learned initially: 10. Number of new classes learned with each succeeding task: 90 for the 2-task case, 15 for the 7-task case, 6 for the 16-task case, and 3 for the 31-task case. Bold indicates the highest value (similarly hereafter).}
 \label{tab:cl2r_cifar}
    \setlength{\tabcolsep}{4pt}
    \centering 
    \sisetup{detect-weight=true,detect-all,table-format=-1.3,round-mode=places}
        \begin{tabular}{l 
                        S[round-precision=2]
                        S[round-precision=2]
                        S[round-precision=2]
                        S[round-precision=2]
                        S[round-precision=2]
                        S[round-precision=2]
                        S[round-precision=2]
                        S[round-precision=2]
                        S[round-precision=2]
                        S[round-precision=2]
                        S[round-precision=2]
                        S[round-precision=2]
                        } 
            \toprule 
            \multirow{4}{*}{\shortstack{\textsc{Method}}} &
            \multicolumn{3}{c}{2 tasks} &
            \multicolumn{3}{c}{7 tasks} &
            \multicolumn{3}{c}{16 tasks} 
            &         \multicolumn{3}{c}{31 tasks}
            \\  
            \cmidrule(lr){2-4}  \cmidrule(lr){5-7} 
            \cmidrule(lr){8-10}  \cmidrule(lr){11-13}  
            
            &
            {$AC$} &
            {$AA$ } &
            {$\mathit{ACA}$ } &
            {$AC$} &
            {$AA$ } &
            {$\mathit{ACA}$ } &
            {$AC$} &
            {$AA$ } &
            {$\mathit{ACA}$ } 
            & {$AC$} &
            {$AA$ } &
            {$\mathit{ACA}$ } 
            \\ 
            \midrule
            \hspace{0pt}ER baseline & 1 & 46.669 & 39.252 & 0.1905 &  46.75 & 7.789 & 0.04167 &  44.905 & 1.69 & 0.01935 & 40.776 & 0.9265 \\
            \hspace{0pt}FAN~\cite{iscen2020memory} & 1  & 48.368  & 43.866 & 0.1429  & 47.587 & 5.631 & 0.15 & 46.098 & 6.691  & 0.07312  & 43.124 & 3.419\\
            \hspace{0pt}{BCT-ER} & 1 & 47.105 & 41.14 & 0.09524  & 46.818 & 3.475 &0& 43.632&0& 0.008602  & 41.964 & 0.4014 \\ 
            \hspace{0pt}LCE-ER & 1  & 49.777 & 50.02 & 0.2381  & 45.147 & 10.383 &  0.2833  & 41.982 & 12.607 &  0.1419  & 38.446 & 5.931 \\ 
            \hspace{0pt}AdvBCT-ER & 1  & 46.277  & 39.712 &0&  35.656 &0& 0.008333  & 34.888 & 0.3899 & 0.02366  & 33.928  & 1.04  \\ 
            \hspace{0pt}CVS~\cite{Wan_2022_CVPR} & 1  & \hspace{-3pt}\textbf{51.31} & \hspace{-3pt}\textbf{52.89} & 0.2857  & \hspace{-3pt}\textbf{51.69} & 12.697 & 0.1333  & \hspace{-3pt}\textbf{49.38} & 5.647 & 0.06882  & 46.33 & 2.928 \\ 
            \hspace{0pt}$d$-Simplex-FD~\cite{biondi2022cl2r} & 1 & 48.762 & 44.958 & 0.38 & 49.67 & 17.31 & 0.45 & 48.42 & 21.32 &  0.34 & 44.822 & 15.37 \\
            \hspace{0pt}$d$-Simplex-HOC (this paper) & 1 & 49.98 & 48.48 & \textbf{0.86} & 48.211 & \hspace{-3pt}\textbf{42.41} & \textbf{0.90} & 47.56 & \hspace{-3pt}\textbf{43.20} & \textbf{0.49} & \hspace{-3pt}\textbf{47.34} & \hspace{-3pt}\textbf{23.44} \\ 
            \bottomrule
        \end{tabular}  
\end{table*}

\begin{table*}[t]
 \caption{1:N search CL$^2$R scenario. Performance metrics with TinyImageNet200/20 for 2, 7, 16, and 31 tasks. Number of classes learned initially: 50. Number of new classes learned with each succeeding task: 150 for the 2-task case, 25 for the 7-task case, 10 for the 16-task case, and 5 for the 31-task case. }
 \label{tab:cl2r_tiny}
    \setlength{\tabcolsep}{4pt}
    \centering 
    \sisetup{detect-weight=true,detect-all,table-format=-1.3,round-mode=places}
        \begin{tabular}{l 
                        S[round-precision=2]
                        S[round-precision=2]
                        S[round-precision=2]
                        S[round-precision=2]
                        S[round-precision=2]
                        S[round-precision=2]
                        S[round-precision=2]
                        S[round-precision=2]
                        S[round-precision=2]
                        S[round-precision=2]
                        S[round-precision=2]
                        S[round-precision=2]
                        } 
            \toprule 
            \multirow{4}{*}{\shortstack{\textsc{Method}}} &
            \multicolumn{3}{c}{2 tasks} &
            \multicolumn{3}{c}{7 tasks} &
            \multicolumn{3}{c}{16 tasks} 
            &         \multicolumn{3}{c}{31 tasks}
            \\  
            \cmidrule(lr){2-4}  \cmidrule(lr){5-7} 
            \cmidrule(lr){8-10}  \cmidrule(lr){11-13}  
            
            &
            {$AC$} &
            {$AA$ } &
            {$\mathit{ACA}$ } &
            {$AC$} &
            {$AA$ } &
            {$\mathit{ACA}$ } &
            {$AC$} &
            {$AA$ } &
            {$\mathit{ACA}$ } 
            & {$AC$} &
            {$AA$ } &
            {$\mathit{ACA}$ } 
            \\ 
            \midrule
            \hspace{0pt}ER baseline &  1 & 34.87 & 34.87 &0& 24.63 &0&0& 21.873 &0 &  \hspace{-3pt}{<0.01} & 19.145 & 0.04889\\
            \hspace{0pt}FAN~\cite{iscen2020memory} & 1 & 33.94 & 33.94 & 0& 33.94 &0&0& 22.939 &0& \hspace{-3pt}{<0.01} & 20.106 & 0.1003  \\
            \hspace{0pt}{BCT-ER} & 1 & 34.67 & 34.67 &0& 24.27 &0& 0.008333 & 22.772 & 0.2206 & \hspace{-3pt}{<0.01} & 19.665 & 0.04409 \\
            \hspace{0pt}LCE-ER  &1 & 34.21 & 34.21 & 0& 22.19 &0& 0& 16.822 &0& 0.02151 & 14.801 & 0.3398 \\    
            \hspace{0pt}AdvBCT-ER  & 1 &  34.32 & 34.32 & 0& 19.77 &0& 0& 19.021 &0& \hspace{-3pt}{<0.01} & 16.612 & 0.04131 \\
            \hspace{0pt}CVS~\cite{Wan_2022_CVPR} & 1 &  \hspace{-3pt}\textbf{37.14} &  \hspace{-3pt}\textbf{37.14} & 0.09524 &  \hspace{-3pt}\textbf{33.20}  & 3.287 &  0.05 &  29.79 & 1.62 & 0.0129 & 24.709 & 0.3504\\
            \hspace{0pt}$d$-Simplex-FD~\cite{biondi2022cl2r} &0& 32.199 &\hspace{-3pt}0&  0.14 & 30.522 &  4.43 &  0.05 & 29.173 & 1.548 &  0.02 &  27.09 & 0.44 \\
            \hspace{0pt}$d$-Simplex-HOC (this paper) & 1 &  36.85 & 36.85  &  \textbf{0.81} & 31.45 & \hspace{-3pt}\textbf{25.96} & \textbf{0.82} &  \hspace{-3pt}\textbf{31.91} & \hspace{-3pt}\textbf{26.72} & \textbf{0.77} & \hspace{-3pt}\textbf{32.35} & \hspace{-3pt}\textbf{25.14} \\ 
            \bottomrule
        \end{tabular}  
\end{table*}

\begin{table*}
 \caption{1:N search CL$^2$R scenario. Performance metrics with CUB180/20 for 2, 7, 16, and 31 tasks. Number of classes learned initially: 60. Number of new classes learned with each succeeding task: 120 for the 2-task case, 20 for the 7-task case, 8 for the 16-task case, and 4 for the 31-task case.}
 \label{tab:cl2r_cub}
    \setlength{\tabcolsep}{4pt}
    \centering 
    \sisetup{detect-weight=true,detect-all,table-format=-1.3,round-mode=places}
        \begin{tabular}{l 
                        S[round-precision=2]
                        S[round-precision=2]
                        S[round-precision=2]
                        S[round-precision=2]
                        S[round-precision=2]
                        S[round-precision=2]
                        S[round-precision=2]
                        S[round-precision=2]
                        S[round-precision=2]
                        S[round-precision=2]
                        S[round-precision=2]
                        S[round-precision=2]
                        } 
            \toprule 
            \multirow{4}{*}{\shortstack{\textsc{Method}}} &
            \multicolumn{3}{c}{2 tasks} &
            \multicolumn{3}{c}{7 tasks} &
            \multicolumn{3}{c}{16 tasks} 
            &         \multicolumn{3}{c}{31 tasks}
            \\  
            \cmidrule(lr){2-4}  \cmidrule(lr){5-7} 
            \cmidrule(lr){8-10}  \cmidrule(lr){11-13}  
            
            &
            {$AC$} &
            {$AA$ } &
            {$\mathit{ACA}$ } &
            {$AC$} &
            {$AA$ } &
            {$\mathit{ACA}$ } &
            {$AC$} &
            {$AA$ } &
            {$\mathit{ACA}$ } 
            & {$AC$} &
            {$AA$ } &
            {$\mathit{ACA}$ } 
            \\ 
            \midrule
            \hspace{0pt}ER baseline & 1 & \hspace{-3pt}\textbf{46.83} & 46.26 & 0.4762 &  43.768 & 20.797 & 0.25 & 42.85 & 10.55 & 0.07312 & 38.536 & 3.076 \\ 
            \hspace{0pt}FAN~\cite{iscen2020memory} & 1 & 45.578 & 43.707  & 0.619 & 43.774 & 26.247 & 0.35 & 41.459 & 14.487 & 0.1011 & 36.647 & 4.163\\ 
            \hspace{0pt}{BCT-ER} & 1 & 44.898 & 43.707  & 0.619 & \hspace{-3pt}\textbf{45.15} & 27.057 & 0.2583 & 41.372 & 11.029 & 0.07527 & 38.264 & 4 \\ 
            \hspace{0pt}LCE-ER  & 1 & 46.54 & \hspace{-3pt}\textbf{47.62} & 0.09524 & 39.614 &  3.912 & 0.09167 & 35.254 & 3.465 & 0.05376 & 30.952 & 1.819  \\    
            \hspace{0pt}AdvBCT-ER  & 1 & 43.367 & 41.667 & 0.381 & 40.549 & 15.452 & 0.125 & 37.953 & 4.76 & 0.06667 & 36.247 &  2.726 \\    
            \hspace{0pt}CVS~\cite{Wan_2022_CVPR}   & 1 & 45.578 & 46.088 &  0.619 & 44.75 & 27.276 & 0.2333 & \hspace{-3pt}\textbf{43.12} & 9.872 & 0.1419 & 39.674 & 5.956 \\
            \hspace{0pt}$d$-Simplex-FD~\cite{biondi2022cl2r} &1 & 39.399 & 36.735 & 0.86 & 39.735 & 35.08 & 0.60 & 39.093 & 24.03 & 0.55 & 39.42 & 21.97 \\
            \hspace{0pt}$d$-Simplex-HOC (this paper) &  1 & 41.61 & 42.517 & \textbf{0.90} & 41.503 & \hspace{-3pt}\textbf{37.72}  & \textbf{0.73} & 41.166 & \hspace{-3pt}\textbf{30.01} & \textbf{0.60} & \hspace{-3pt}\textbf{42.01} & \hspace{-3pt}\textbf{25.46} \\
            \bottomrule
        \end{tabular}  
\end{table*}

\begin{figure*}[t]
    \centering
    \hspace{-7pt}
    \begin{subfigure}{0.245\linewidth}
        \centering
        \includegraphics[width=0.95\linewidth]{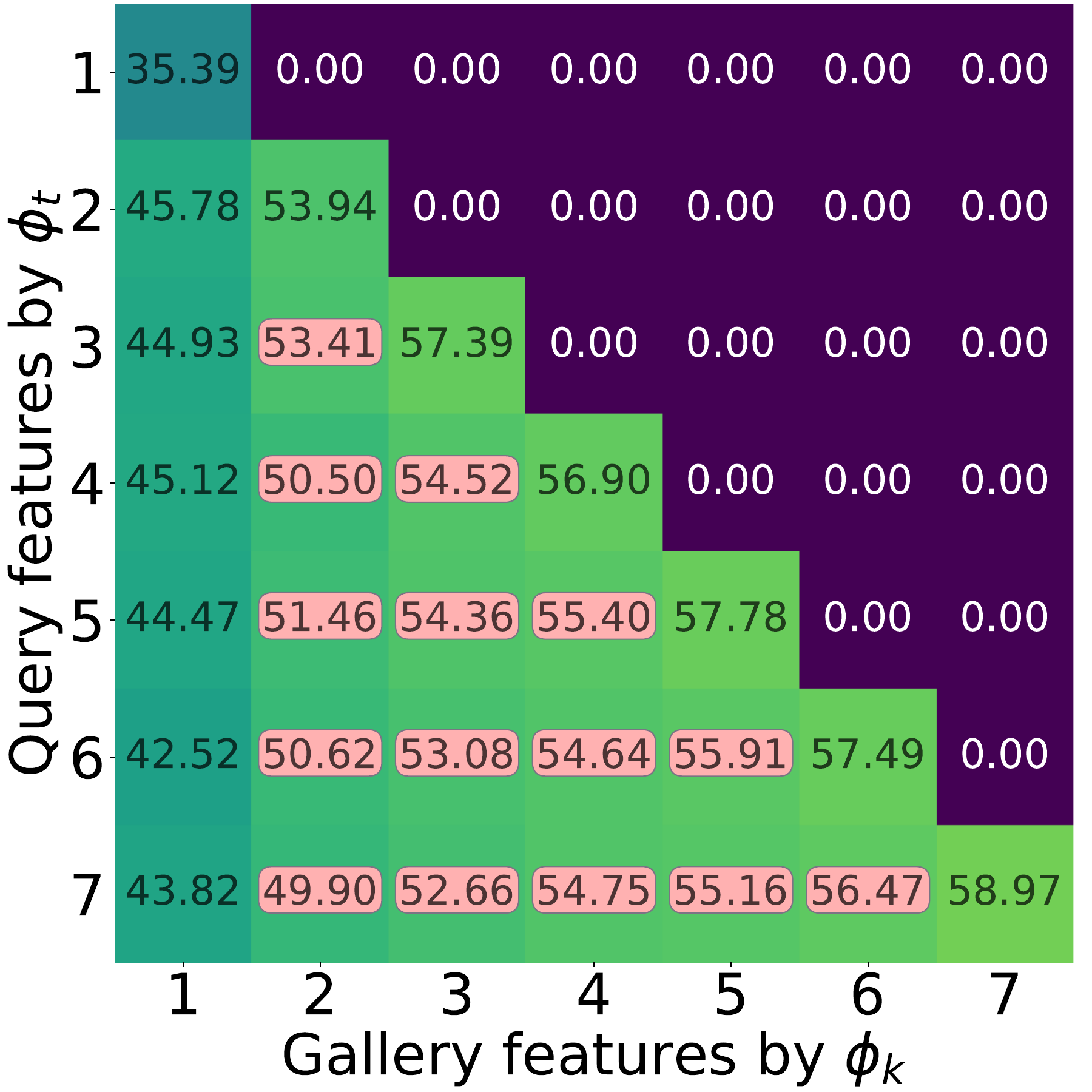}
        \caption{CVS~\cite{Wan_2022_CVPR}}
    \label{fig:fan-10cifar-cl2r}
    \end{subfigure}
    \hspace{-10pt}
    \begin{subfigure}{0.245\linewidth}
        \centering
        \includegraphics[width=0.95\linewidth]{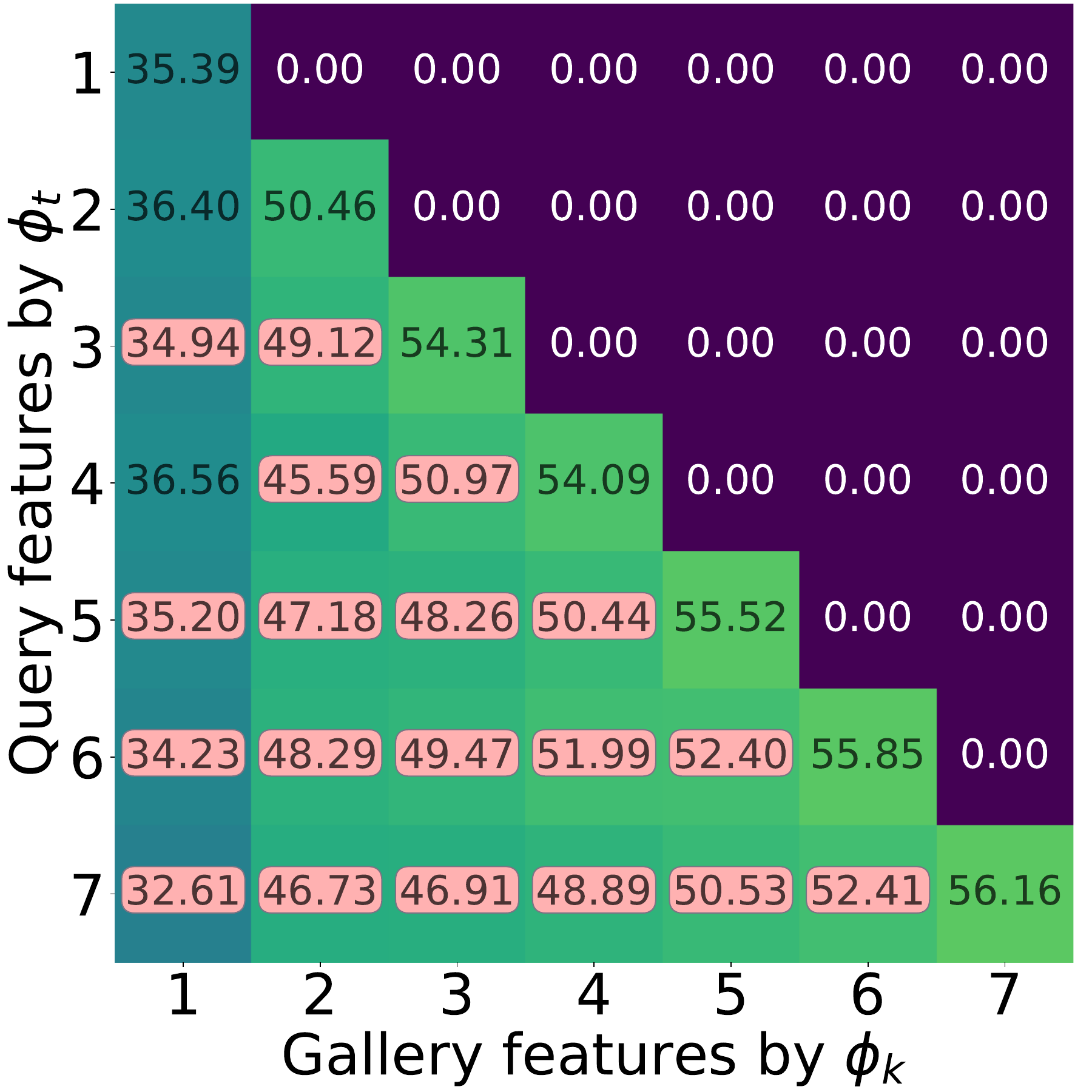}
        \caption{BCT-ER~\cite{shen2020towards}}
    \label{fig:bct-10cifar-cl2r}
    \end{subfigure}
    \hspace{-10pt}
    \begin{subfigure}{0.245\linewidth}
        \centering
        \includegraphics[width=0.95\linewidth]{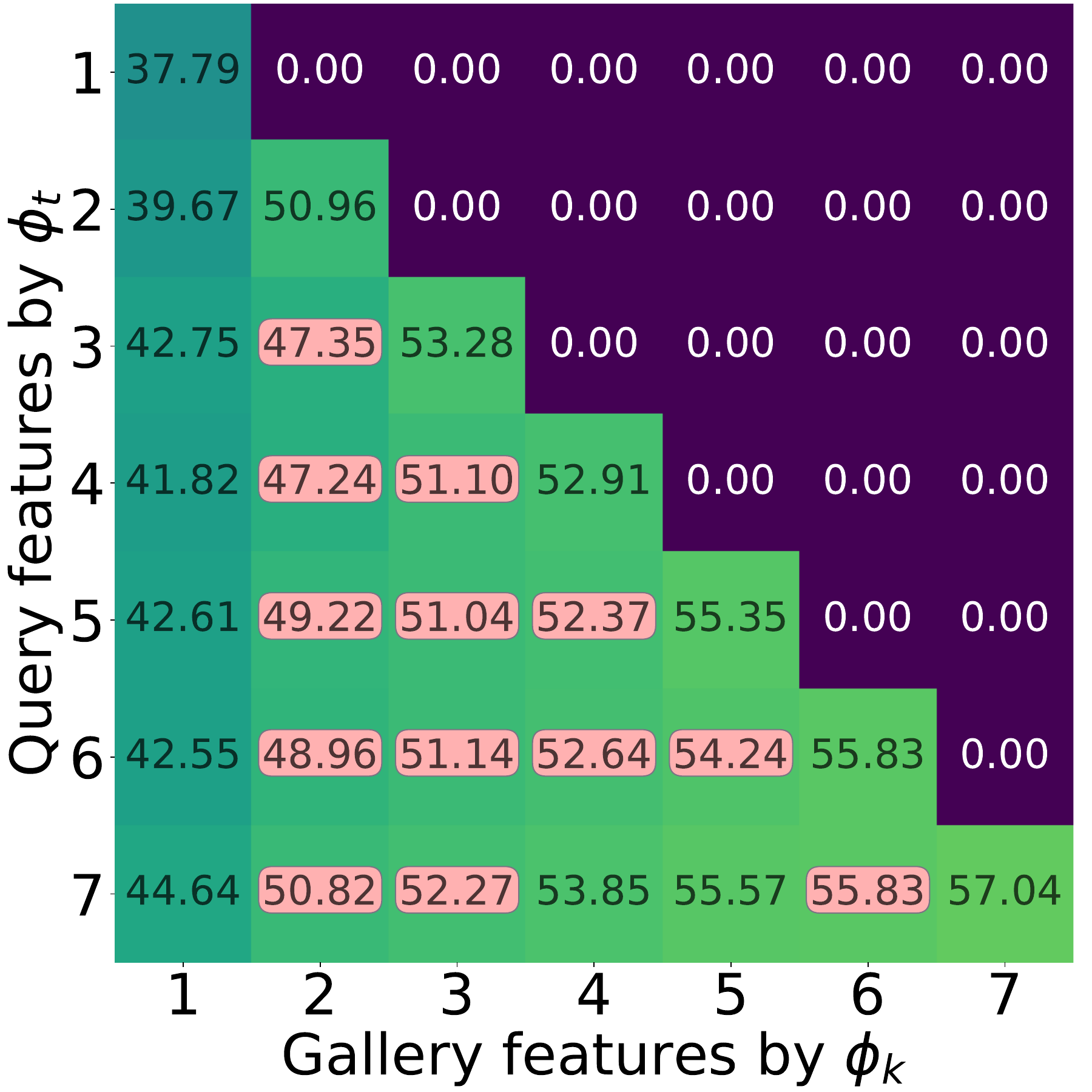}
        \caption{$d$-Simplex-FD~\cite{biondi2022cl2r}}
    \label{fig:simplex-fd-10cifar-cl2r}
    \end{subfigure}
    \hspace{-10pt}
    \begin{subfigure}{0.245\linewidth}
        \centering
        \includegraphics[width=0.95\linewidth]{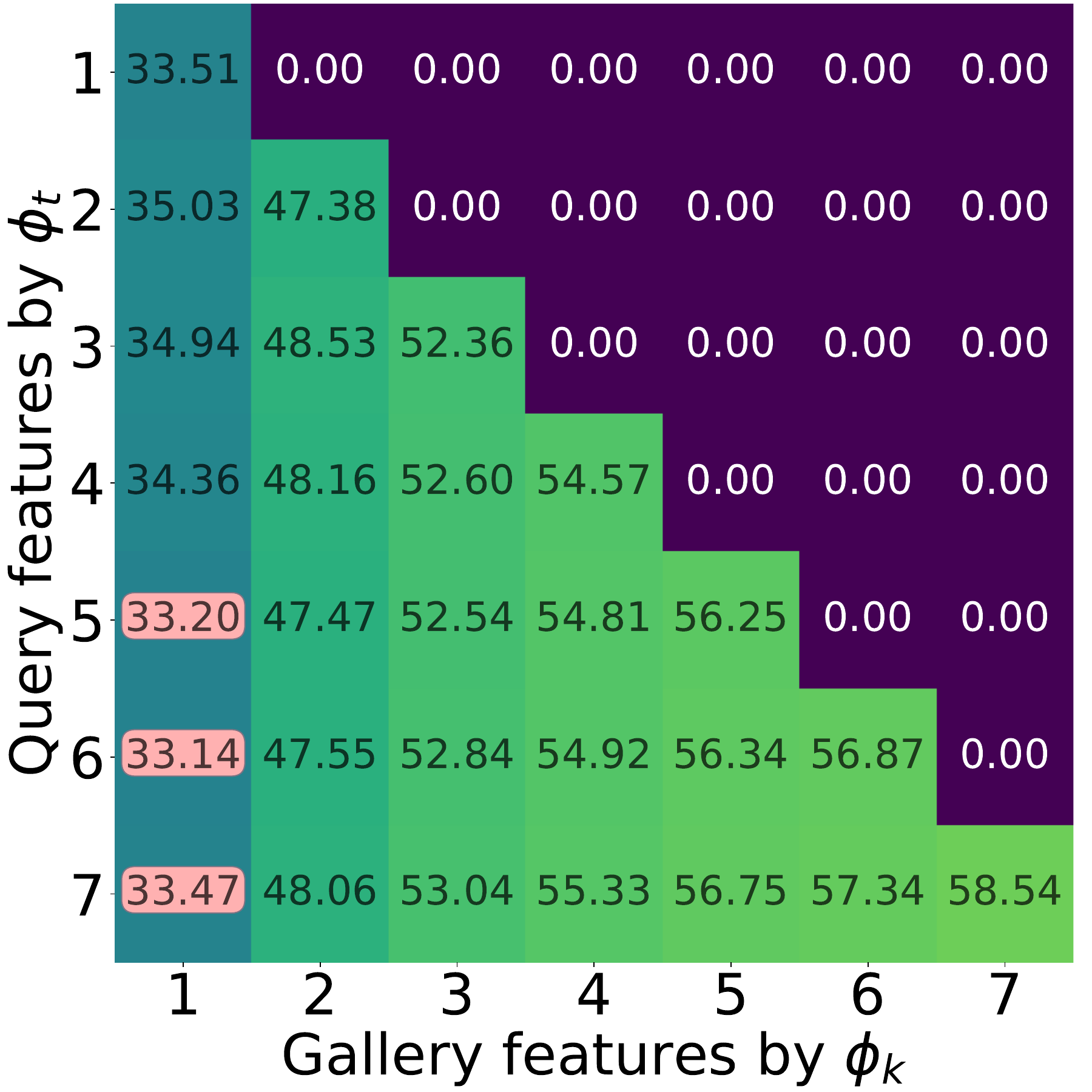}
        \caption{$d$-Simplex-HOC (this paper)}
    \label{fig:our-10cifar-cl2r}
    \end{subfigure}
    \hspace{-16pt}
    \begin{subfigure}{0.08\linewidth}
        \centering
        \includegraphics[width=0.385\linewidth]{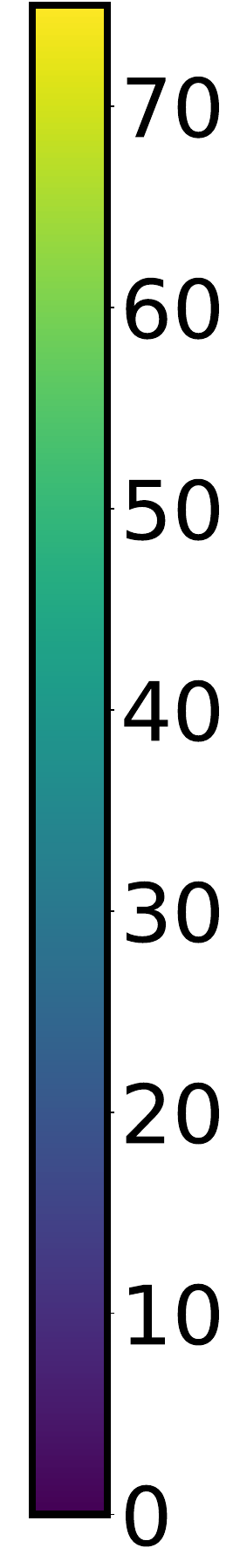}        
        \vspace{1cm}
    \end{subfigure}
    \caption{
    1:N search CL$^2$R scenario. Compatibility Matrices of CVS, BCT-ER, $d$-Simplex-FD, and $d$-Simplex-HOC on CIFAR100/10 for 7 tasks. Entries that do not satisfy Eq.~\ref{eq:multistepecc} are highlighted with light-red background.}  
    \label{fig:cifar-10cl2r-compmatrix}
\end{figure*}

\subsection{Comparative analysis for the CL\texorpdfstring{$^2$}{2}R scenario} \label{sec:experiments_cl2r}

For the CL$^2$R scenario, we consider distinct settings as detailed in Tab.~\ref{tab:cl2r_settings}, including datasets, network architectures, feature dimensions, and training hyper-parameters. 

The number of classes used in the initial task is 10 for CIFAR100, 50 for TinyImageNet200, and 60 for CUB180. The remaining classes of the training datasets are split into multiple task-sets of equal size and used for fine-tuning (see captions in Tabs.~\ref{tab:cl2r_cifar},~\ref{tab:cl2r_tiny},~\ref{tab:cl2r_cub}). 
The Experience Replay buffer is set to 20 images per class.

Tab.~\ref{tab:cl2r_cifar}, Tab.~\ref{tab:cl2r_tiny}, and Tab.~\ref{tab:cl2r_cub} report $AC$, $AA$, and $\mathit{ACA}$ values using CIFAR100/10, TinyImageNet200/20, and CUB180/20 datasets, respectively.
The results indicate that stationarity learned through the $d$-Simplex fixed classifier is crucial for achieving compatibility.
With the exception of the $d$-Simplex-FD and $d$-Simplex-HOC methods, we observed a consistent decline in performance  across all experiments as the number of tasks increases, reaching a significant drop at 31 tasks. 

Overall, $d$-Simplex-HOC outperforms other approaches in terms of compatibility ($AC$ and $\mathit{ACA}$), with high accuracy ($AA$) in the experiments over all the datasets, particularly in the experiment involving 31 tasks. 
It is important to highlight the significantly high $\mathit{ACA}$ score, which indicates the ability of $d$-Simplex-HOC to achieve the best trade-off between compatibility and accuracy.
The results also show that, by taking into account higher-order dependencies between previous and current representations, the $d$-Simplex-HOC succeeds to effectively capture and utilize the accumulated knowledge from previous tasks. This holds across all datasets, regardless of their diversity. 

Among the other methods, CVS achieves stationarity indirectly by enforcing constraints that align the current model with previous models, thereby maintaining consistency between learned representations. 
It constrains on the model output while enforcing coherence with with replayed embeddings from previous tasks. Nevertheless, compatibility drops as the number of tasks increases.  
Differently, FAN achieves compatibility by employing mappings between representations learned from different tasks. This approach, however, has drawbacks: it increases computational cost and reduces performance, since ensuring compatibility between non-sequential representations requires composing multiple mapping functions.
All other methods show reduced performance in all experiments except in the case in which the model is subjected to a single update.

Further performance details are shown in Fig.~\ref{fig:cifar-10cl2r-compmatrix}, which reports the Compatibility Matrices of CVS, BCT-ER, $d$-Simplex-FD, and $d$-Simplex-HOC for the CIFAR100/10 dataset with 7 tasks.  
We observe that $d$-Simplex-HOC scores both the highest number of times compatibility is assessed and the highest accuracy values. 
The contribution of $\mathcal{L}_\text{\loss}$~loss with respect to the $d$-Simplex cross-entropy loss with the mean square error loss, as in \mbox{$d$-Simplex-FD}, is evident when comparing the Compatibility Matrices of $d$-Simplex-HOC and $d$-Simplex-FD methods (respectively, (d) and (c) in the figure).

\begin{figure}[t]
    \centering
    \vspace{-0.1cm}
    \includegraphics[width=\linewidth]{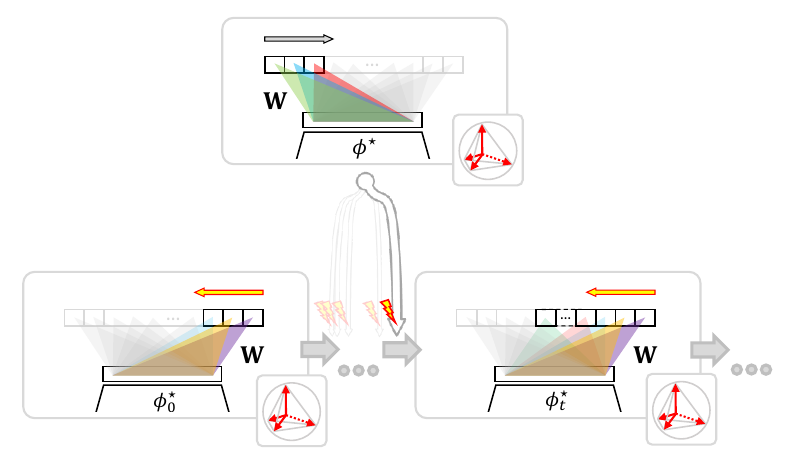}
    \vspace{-0.6cm}
    \caption{Fine-tuning and model replacement in the IAM-CL$^2$R scenario. Class labels of the pre-trained model $\phi^\star$ are assigned from left to right. 
    When fine-tuning, class labels are assigned from right to left. 
    This results in distinct class labels for pre-training and fine-tuning. The fixed matrix $\mathbf{W}$ of the $d$-Simplex fixed classifier establishes a common interface between the models.}
    \label{fig:iamcl2r-approach}
\end{figure}

\subsection{Comparative analysis for the IAM-CL\texorpdfstring{$^2$}{2}R scenario} \label{sec:experiments_iamcl2r}

In the IAM-CL$^2$R scenario, we explore two types of model replacement: (a) where a fine-tuned model is replaced with an improved model trained from scratch on additional data; (b) where the fine-tuned model is replaced with a model that uses a more powerful architecture.

\begin{table}[t]
\centering
\caption{Settings for the reduced-scale (CIFAR100R/10) and large-scale (CelebA) IAM-CL$^2$R scenarios.}
\label{tab:experimental_settings}
\setlength{\tabcolsep}{2pt}
\begin{tabular}{@{}lcc@{}}
\toprule
\textsc{Settings} & CIFAR100R/10 & CelebA \\
\midrule
\multicolumn{3}{l}{\textit{Datasets}} \\
Pre-training Dataset & ImageNet32 \cite{chrabaszcz2017downsampled} & CASIA-WebFace \cite{casiawebface} \\
\quad - n. classes & 100 & 3525 \\
Fine-tuning Dataset & CIFAR100R*  & CelebA \cite{liu2015faceattributes} \\
\quad - n. classes & 100 & 8192 \\
Test Dataset & CIFAR10 & CelebA test split \\
\midrule
\multicolumn{3}{l}{\textit{Network Architecture}} \\
Backbone & ResNet18 & DINOv3 ViT-B16 \cite{simeoni2025dinov3} \\
Input Size & 32$\times$32 & 224$\times$224 \\
\midrule
\multicolumn{3}{l}{\textit{Feature Dimensions}} \\
$d$-Simplex ($K-1$) & 1023 ($K$=1024) & 21999 ($K$=22000) \\
Compared Methods** & 512 & 768 \\
\midrule
\multicolumn{3}{l}{\textit{Fine-tuning Details}} \\
Fine-tuning tasks & 7, 31 & 16 \\
Classes per task & (10 + 6$\times$15), (10 + 30$\times$3) & 16$\times$512 \\
\midrule
\multicolumn{3}{l}{\textit{Replacement Model}} \\
Replacements at & 40, 70 classes & 2560, 5120 classes \\
Training Dataset & ImageNet32 & CASIA-WebFace \\
\quad - n. classes & 300, 600 & 7050, 10575 \\
\midrule
\multicolumn{3}{l}{\textit{Pre-training Hyperparameters}} \\
Optimizer & SGD & AdamW \\
Epochs & 300 & 90 \\
Batch Size & 128 & 320 \\
Initial LR & 0.1 & 0.001 \\
LR Schedule & Cosine Annealing & Cosine Annealing \\
\midrule
\multicolumn{3}{l}{\textit{Fine-tuning Hyperparameters}} \\
Optimizer & SGD & AdamW \\
Epochs & 70 & 90 \\
Batch Size & 128 & 320 \\
Initial LR & 0.001 & 0.001 \\
LR Decay & div. by 10 after 50, 64 epochs & Cosine Annealing \\
Experience Replay & 20 images/class & 10 images/class \\
\bottomrule
\end{tabular}
\begin{flushleft}
* CIFAR100R includes 100 CIFAR100 classes with 300 images per class. \\
** The feature size corresponds to the size of the penultimate layer of the backbone.
\end{flushleft}
\end{table}

\begin{table}[t]
    \captionof{table}{1:N search IAM-CL$^2$R scenario. Performance metrics with CIFAR100R/10 for 7 tasks and 31 tasks with two model replacements. 
    The improved models were obtained by retraining ResNet18 from scratch with 300 and 600 ImageNet32 classes.}
    \label{tab:iamcl2r_25tasks}
\centering 
\sisetup{detect-weight=true,detect-all,table-format=-1.3,round-mode=places, round-precision=2}
    \begin{tabular}{l SSSS
                    } 
        \toprule 
        \multirow{4}{*}{\shortstack{\textsc{Method}}}
        & \multicolumn{2}{c}{\textcolor{black}{7 tasks}} & \multicolumn{2}{c}{\textcolor{black}{31 tasks}} \\
        \cmidrule(lr){2-3}  \cmidrule(lr){4-5} 
        
        &
        {$AC$} &
        {$AA$ } &
        {$AC$} &
        {$AA$ }
        \\ 
        \midrule   
    ER baseline &0& 36.217 & \hspace{-3pt}{<0.01} & 31.299 \\
    FAN~\cite{iscen2020memory}  &0& 36.316 & \hspace{-3pt}{<0.01}  & 30.787 \\
    BCT-ER~\cite{shen2020towards}  & 0 & 35.59 &0& 29.878  \\ 
    LCE-ER~\cite{meng2021learning}  &0& 34.891&0& 29.299 \\ 
    AdvBCT-ER~\cite{pan2023boundary} & 0& 35.732 &0& 30.097 \\
    CVS~\cite{Wan_2022_CVPR}  &  0 & 36.306 & 0.006452 & 31.344 \\
    $d$-Simplex-FD~\cite{biondi2022cl2r} & 0.04762 & 56.581  & 0.2108 & 56.267 \\
    $d$-Simplex-HOC (this paper) & \textbf{0.95} & \hspace{-3pt}\textbf{68.13} &  \textbf{0.65} & \hspace{-3pt}\textbf{67.40} \\
        
        \bottomrule
    \end{tabular}  
\end{table}

\begin{figure*}[t]
    \centering
    \begin{subfigure}{\linewidth}
        \centering
        \includegraphics[width=0.65\linewidth]{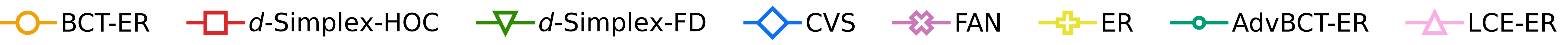}
    \end{subfigure}
    \hspace{-20pt}
    \begin{subfigure}{0.49\linewidth}
        \centering
        \includegraphics[width=1.03\linewidth]{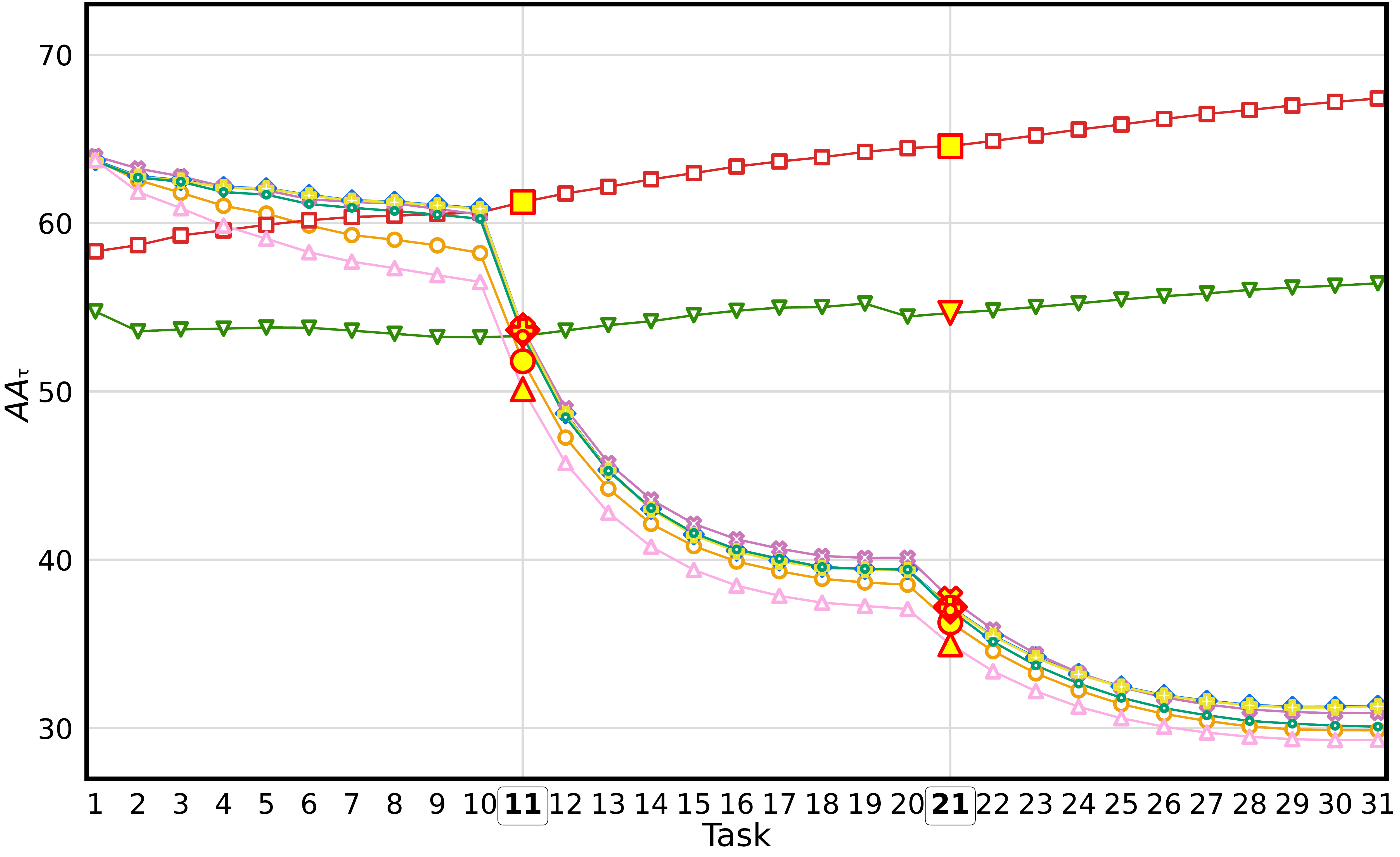} \caption{Two model replacements}\label{fig:cifar30-iamcl2r}
    \end{subfigure}
    \hspace{2pt}
    \begin{subfigure}{0.49\linewidth}
        \centering
        \includegraphics[width=1.03\linewidth]{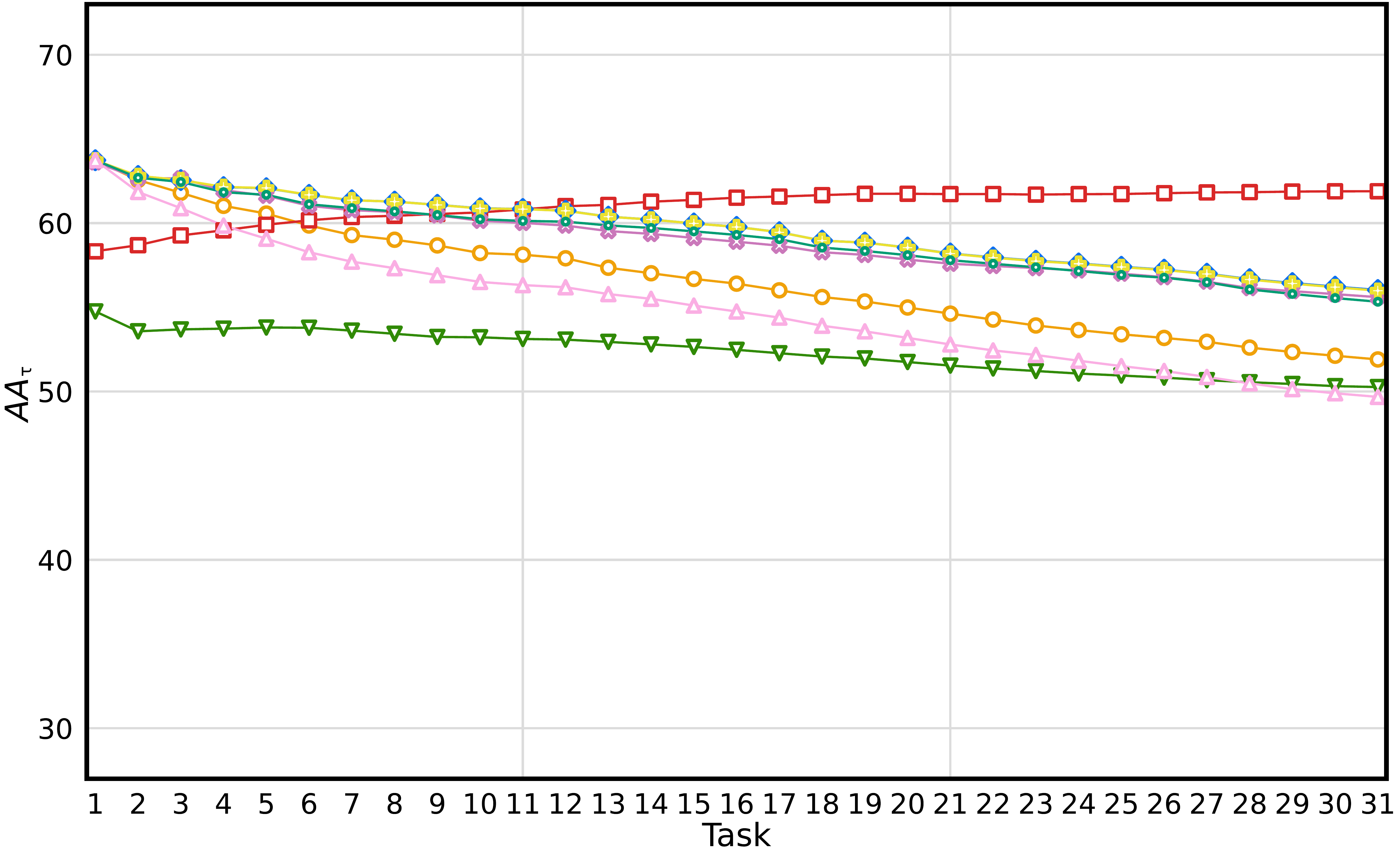} \caption{No model replacements}\label{fig:cifar30-cl2r}
    \end{subfigure}
    \caption{
    1:N search IAM-CL$^2$R scenario. Average Accuracy up to task $\tau$ ($AA_{\tau}$) 
    with CIFAR100R/10 for 31 tasks: (a) two model replacements, where the first tasks after replacements (11 and 21) are marked by boxed task indices; (b) no model replacements. The improved models are obtained by retraining ResNet18 from scratch with 300 and 600 ImageNet32 classes.}
    \label{fig:iamcl2r}
\end{figure*}
\begin{figure*}[t]
    \centering
    \hspace{-7pt}
    \begin{subfigure}{0.245\linewidth}
        \centering
        \includegraphics[width=0.95\linewidth]{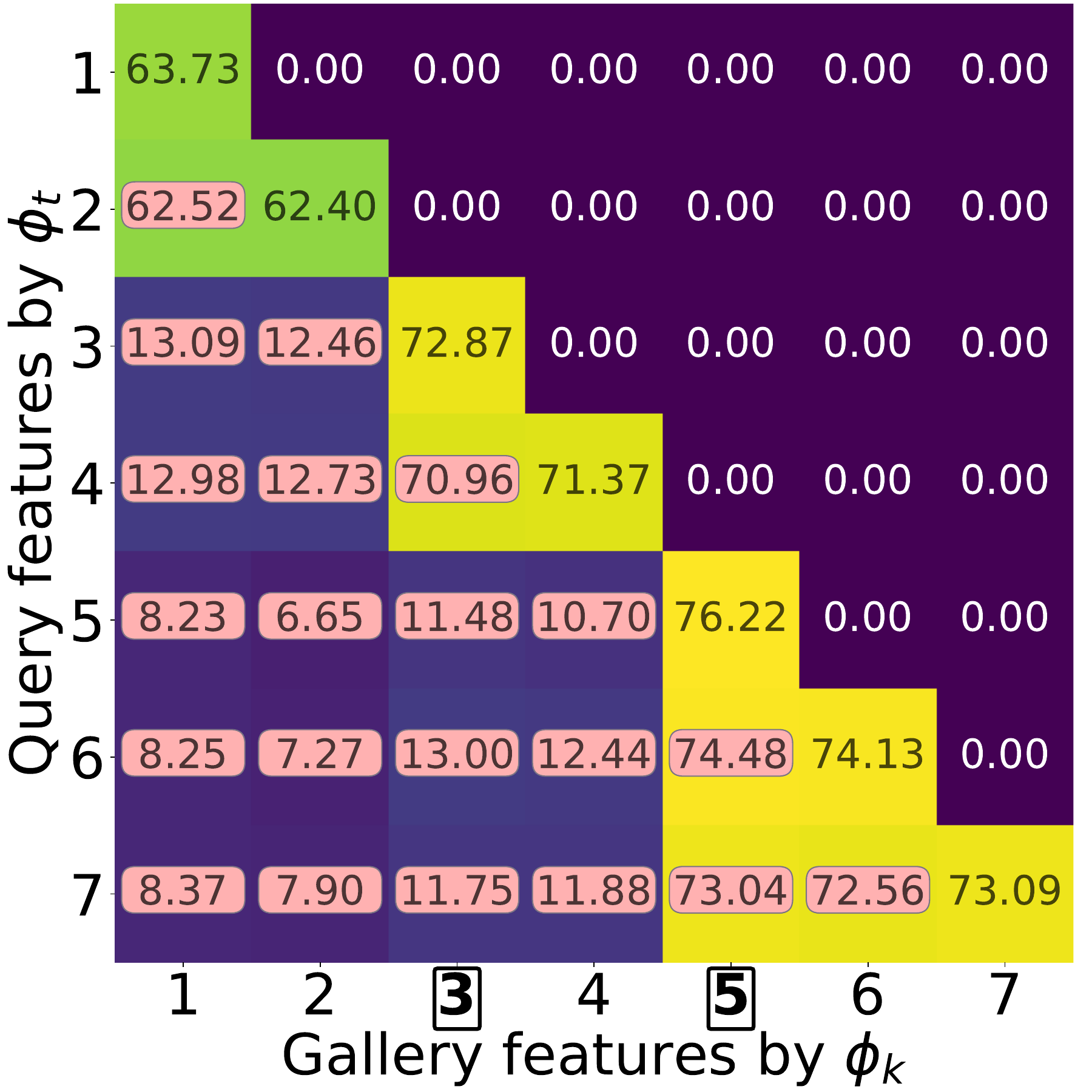}
        \caption{CVS~\cite{Wan_2022_CVPR}}
    \label{fig:fan-10cifar-iamcl2r}
    \end{subfigure}
    \hspace{-10pt}
    \begin{subfigure}{0.245\linewidth}
        \centering
        \includegraphics[width=0.95\linewidth]{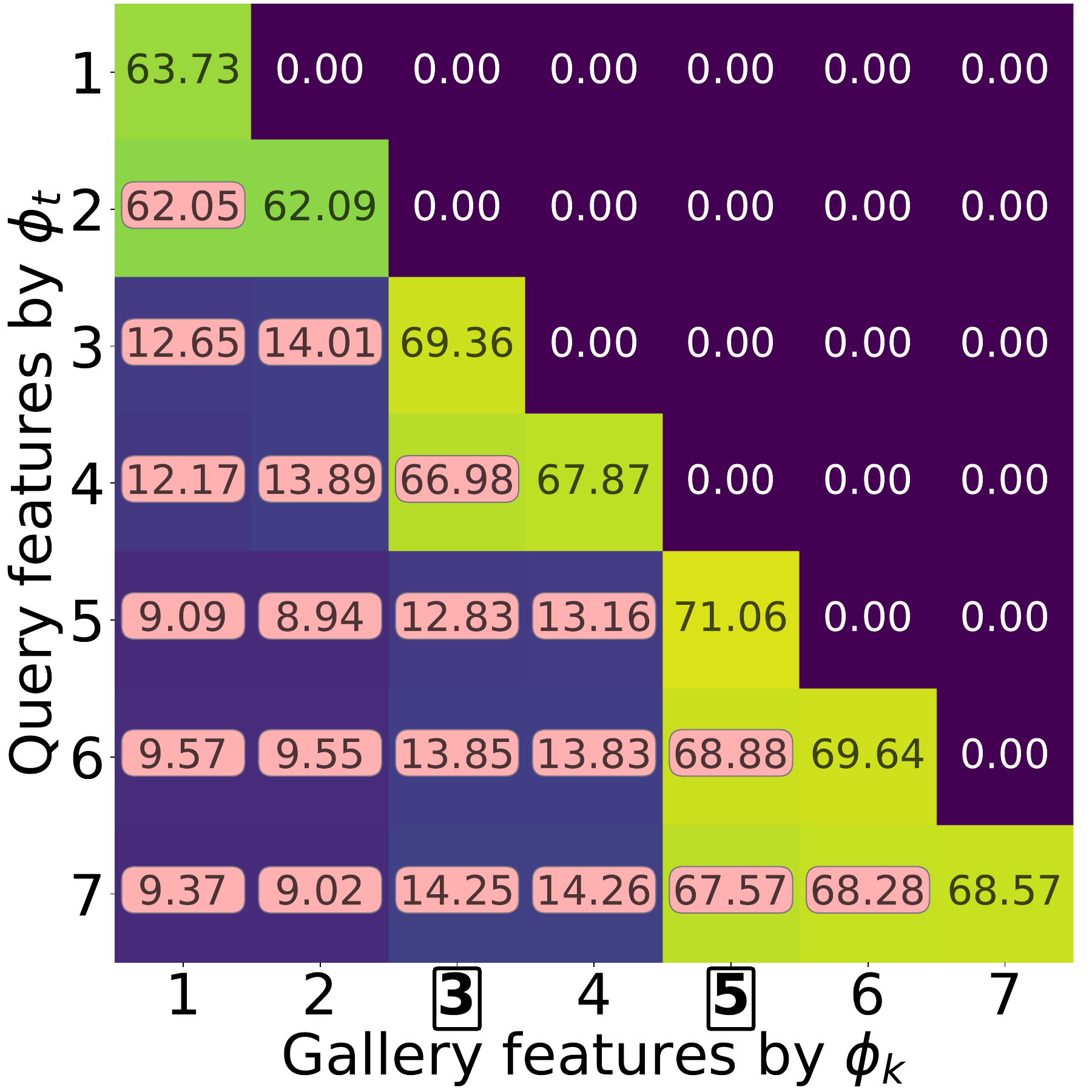}
        \caption{BCT-ER~\cite{shen2020towards}}
    \label{fig:bct-10cifar-iamcl2r}
    \end{subfigure}
    \hspace{-10pt}
    \begin{subfigure}{0.245\linewidth}
        \centering
        \includegraphics[width=0.95\linewidth]{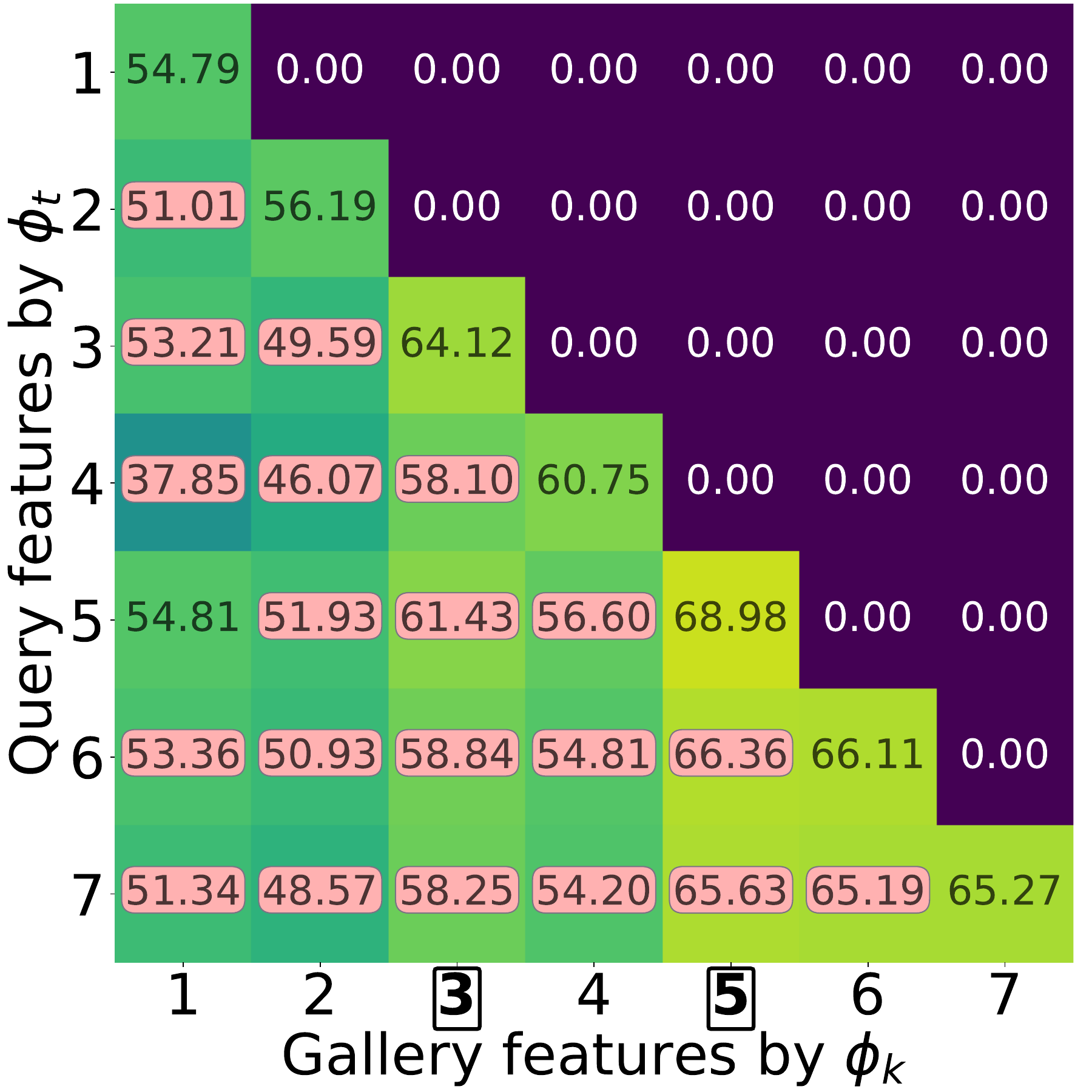}
        \caption{$d$-Simplex-FD~\cite{biondi2022cl2r}}
    \label{fig:simplex-fd-10cifar-iamcl2r}
    \end{subfigure}
    \hspace{-10pt}
    \begin{subfigure}{0.245\linewidth}
        \centering
        \includegraphics[width=0.95\linewidth]{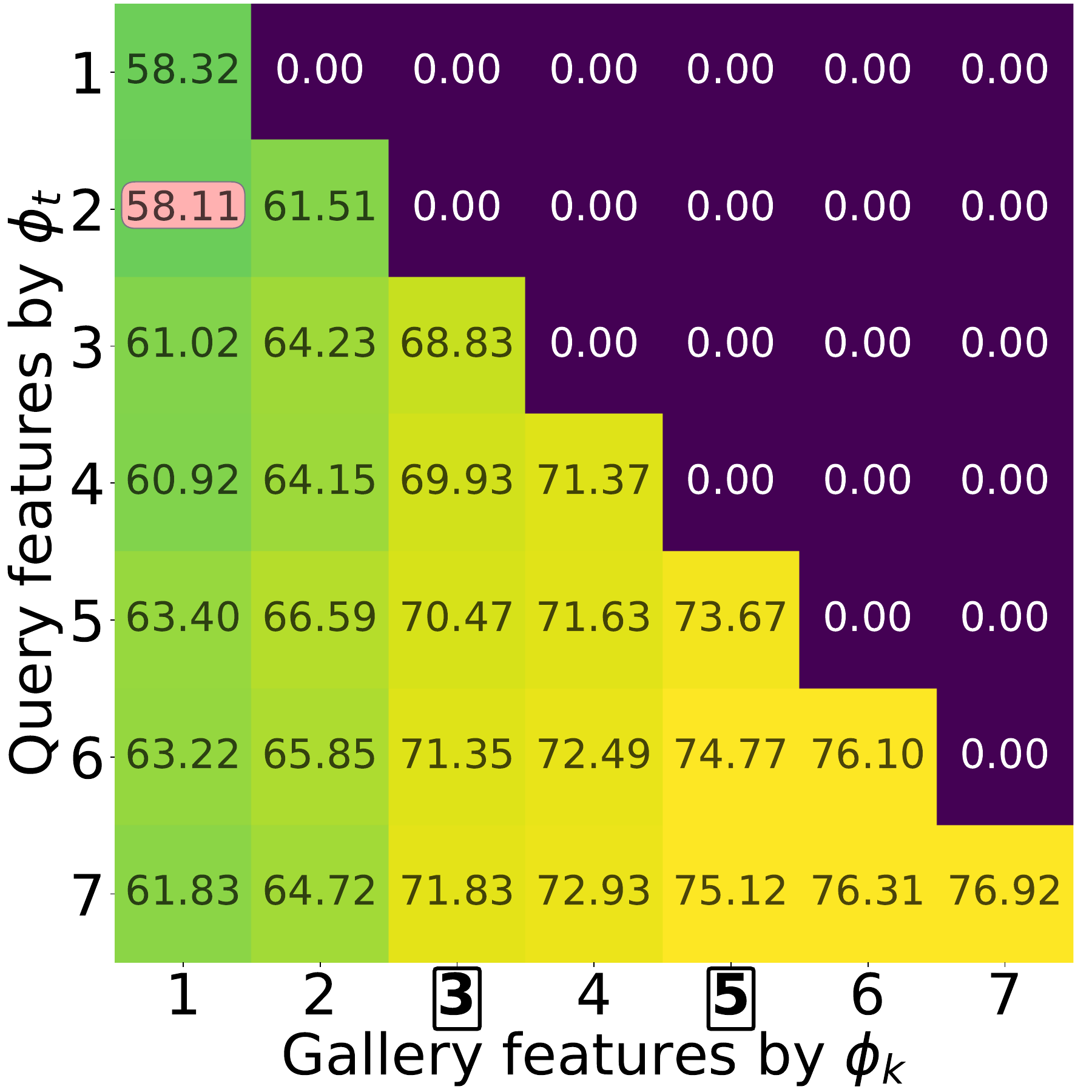}
        \caption{$d$-Simplex-HOC (this paper)}
    \label{fig:our-10cifar-iamcl2r}
    \end{subfigure}
    \hspace{-16pt}
    \begin{subfigure}{0.08\linewidth}
        \centering
        \includegraphics[width=0.385\linewidth]{images/cifar/BAR.pdf}        
        \vspace{1cm}
    \end{subfigure}
    \caption{1:N search IAM-CL$^2$R scenario. Compatibility Matrices of $d$-Simplex-HOC, CVS, BCT-ER, and $d$-Simplex-FD on CIFAR100R/10 for 7 tasks and two model replacements, where the first tasks after replacements (3 and 5) are marked by boxed task indices.
    The improved models were obtained by retraining ResNet18 from scratch with 300 and 600 ImageNet32 classes.
    Entries that do not satisfy compatibility Eq.~\ref{eq:multistepecc} are highlighted with light-red background. 
    }
    \label{fig:cifar-10cl2r-iamcl2r}
\end{figure*}

\begin{figure*}[t]
    \centering
    \begin{subfigure}{\linewidth}
        \centering
        \includegraphics[width=0.55\linewidth]{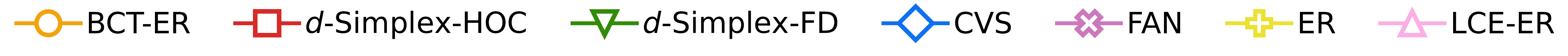}
    \end{subfigure}
    \hspace{-20pt}
    \begin{subfigure}{0.49\linewidth}
        \centering
        \includegraphics[width=1.03\linewidth]{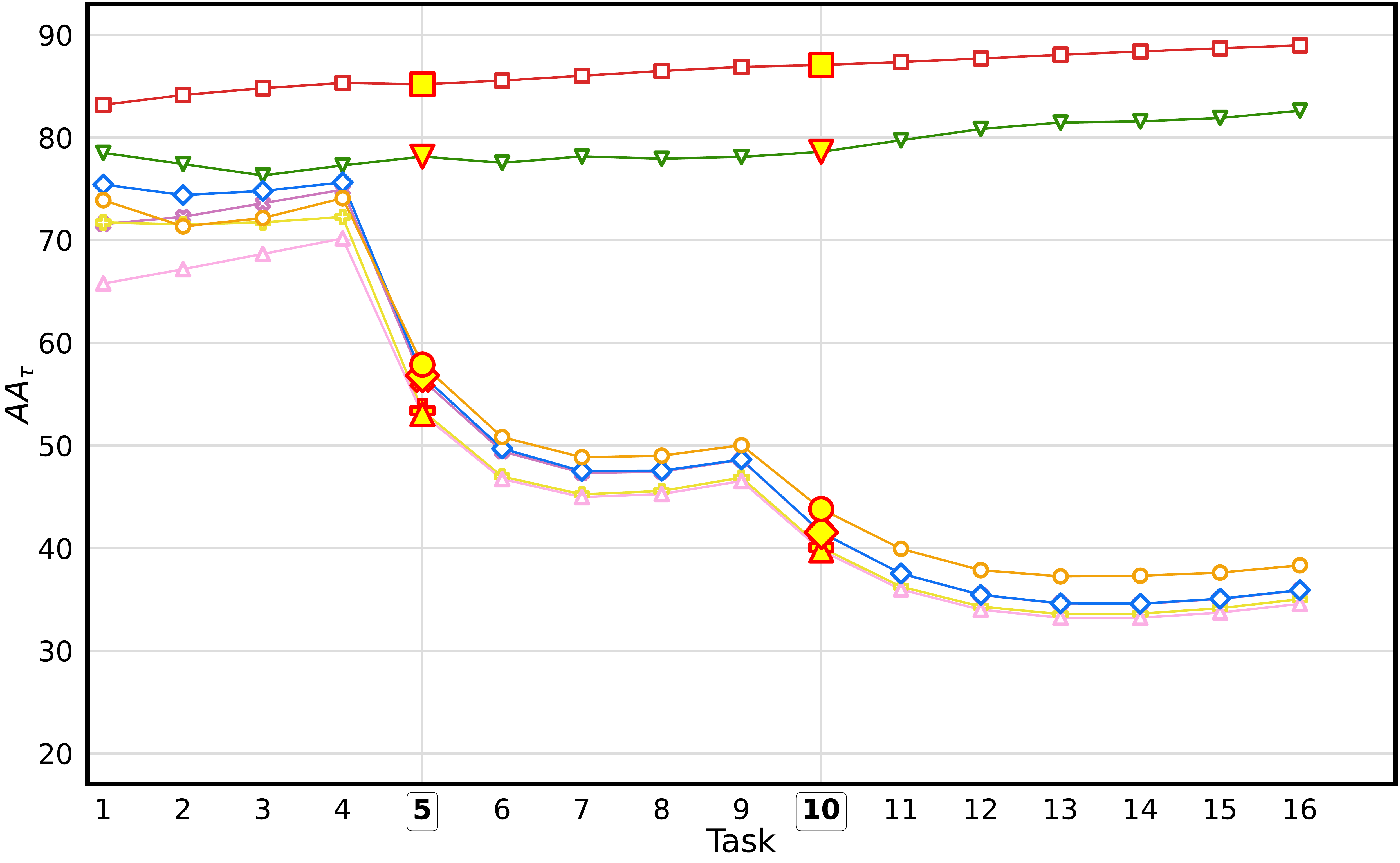} \caption{Two replacements}\label{fig:celeba-iamcl2r}
    \end{subfigure}
    \hspace{2pt}
    \begin{subfigure}{0.49\linewidth}
        \centering
        \includegraphics[width=1.03\linewidth]{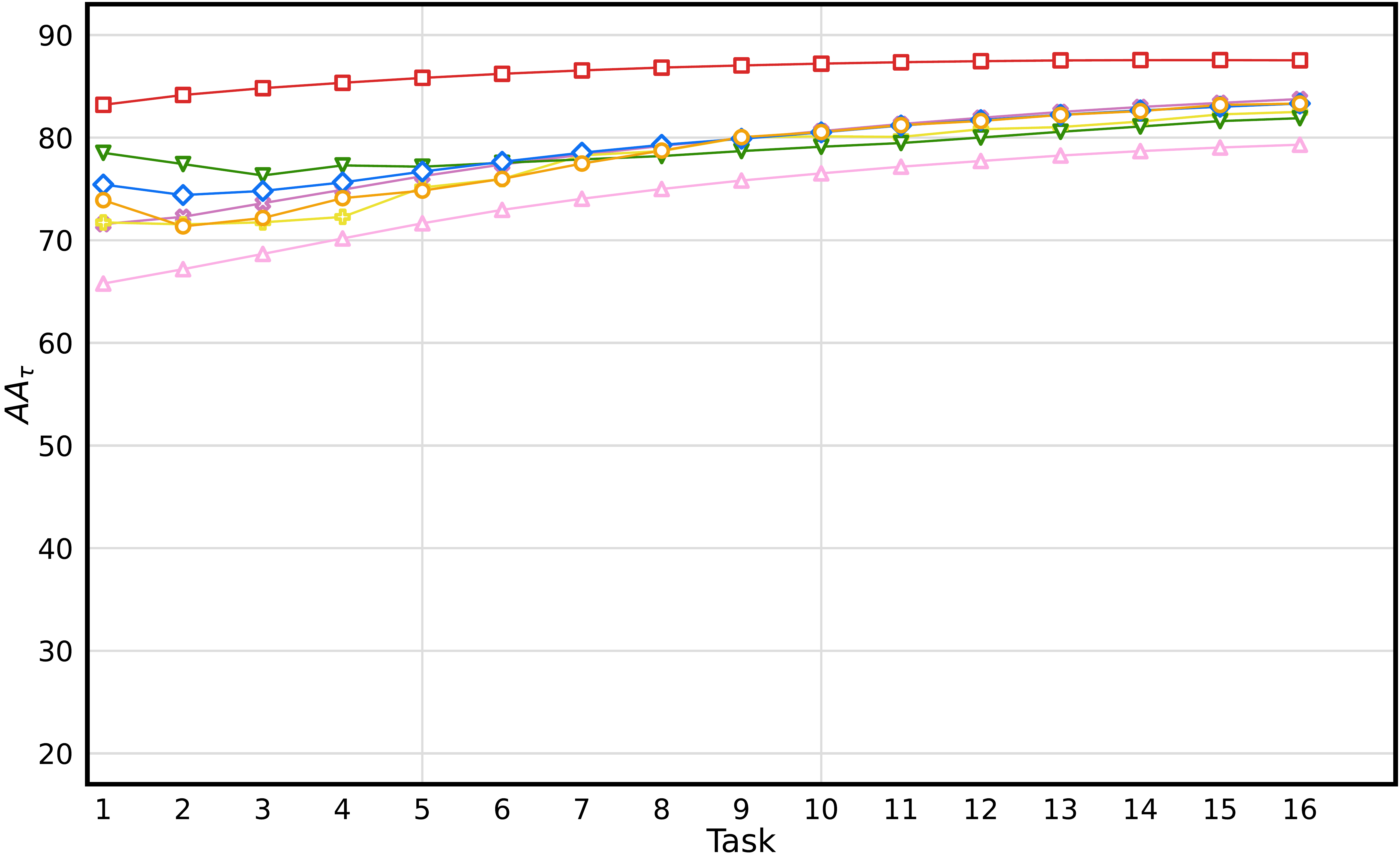} \caption{No replacements}\label{fig:celeba-cl2r}
    \end{subfigure}
    \caption{1:N search IAM-CL$^2$R scenario. Average Accuracy up to task $\tau$ ($AA_{\tau}$) 
    with CelebA for 16 tasks: (a) two model replacements, where the first tasks after replacements (5 and 10) are marked by boxed task indices; (b) no model replacements. The improved models are obtained by fine-tuning a DINOv3 ViT-B/16 with 7050 and 10575 CASIA-WebFace classes, respectively.
    }
    \label{fig:iamcl2r-celeba}
\end{figure*}

For \mbox{$d$-Simplex-HOC} and \mbox{$d$-Simplex-FD}, the use of $d$-Simplex fixed classifier with class pre-allocation allows seamless model replacement. To avoid overlapping between classes assigned for pre-training and for fine-tuning, we have found convenient to assign the class labels of the pre-trained models from left to right and class labels of the fine-tuned models from right to left (see Fig.~\ref{fig:iamcl2r-approach}). 

\noindent\textbf{Model Replacement with Improved Models. \;} 
We analyze this case at two distinct scales, respectively, reduced- and large-scale.
Tab. \ref{tab:experimental_settings} shows the setting details for the reduced- and large-scale IAM-CL$^2$R scenarios.

For the reduced-scale IAM-CL$^2$R scenario, we report experiments using the CIFAR100R/10 datasets with ResNet18 as network architecture.
The improved models are obtained by retraining the ResNet18 network from scratch, using 300 and 600 ImageNet32 classes, respectively.
Tab.~\ref{tab:iamcl2r_25tasks} reports the $AC$ and $AA$ of the methods compared when fine-tuning with 7 tasks (introducing 10 CIFAR100R classes in the first task and 15 classes in each subsequent task) and 31 tasks (introducing 10 CIFAR100R classes in the first task and 3 classes in each subsequent task) and two model replacements. We observe that in both cases, all the models except $d$-Simplex-HOC are not able to achieve significant performance scores. The $d$-Simplex-HOC achieves the highest average compatibility and accuracy.

Fig.~\ref{fig:iamcl2r} shows plots of \mbox{$AA_{\tau}$} with and without model replacement in 31 tasks (Fig.~\ref{fig:cifar30-iamcl2r} and Fig.~\ref{fig:cifar30-cl2r}, respectively). 
The comparison provides valuable insights into how effectively the methods leverage the improved representation. 
Both $d$-Simplex-HOC and $d$-Simplex-FD succeed to incorporate the model improvements and achieve higher average accuracy than in the case with no replacement. 
All the other methods present an evident decay of average accuracy after the replacements and end up with a worse performance. 
This is due to the fact that the fine-tuned model has been obtained by retraining from scratch, resulting in a different representation than the one before the replacement.

Additional performance details can be observed in Fig.~\ref{fig:cifar-10cl2r-iamcl2r}. It shows the Compatibility Matrices of CVS, BCT-ER, $d$-Simplex-FD and $d$-Simplex-HOC with 7 tasks. Model replacements were performed at tasks 3 and 5. The values of self-test and cross-test accuracy highlight that only the $d$-Simplex-based methods are able to leverage the improved expressive power of the models after replacements. Both CVS and BCT-ER score very low cross-tests accuracy across model replacements. This results in representations that are not compatible in most cases.
$d$-Simplex-HOC is the only method that is able to leverage the additional knowledge provided by the replaced models while maintaining high compatibility.
It is worth noting that, although $d$-Simplex-FD also achieves high accuracy values, $d$-Simplex-HOC demonstrates superior compatibility performance. This is due to the high-order alignment of the learned representations, which cannot be achieved by the feature distillation loss in $d$-Simplex-FD.

For the large-scale IAM-CL$^2$R scenario, we report experiments on the CelebA dataset with the DINOv3 ViT-B/16 model.
The initial model---fine-tuned on 3525 CASIA-WebFace classes---is subsequently fine-tuned on CelebA across 16 tasks, with 512 new classes at each task. 
The improved models are obtained by fine-tuning a DINOv3 ViT-B/16 with 7050 and 10575 CASIA-WebFace classes, respectively.

Fig.~\ref{fig:celeba-iamcl2r} shows $AA_\tau$ with two model replacements, and Fig.~\ref{fig:celeba-cl2r} shows the corresponding results without model replacement. 
The results mirror those from the CIFAR100R/10 experiments. 
Both $d$-Simplex-HOC and $d$-Simplex-FD benefit from the improved models, exhibiting  accuracy gains after each replacement, whereas the other methods’ performance drops after replacements.
Notably, $d$-Simplex-HOC achieves the highest performance also in this larger and more complex scenario.\\

\begin{table}[t]
    \centering
    \caption{1:N search IAM-CL$^2$R scenario. Average Compatibility and Average Accuracy values with CIFAR100R/10 for 31 tasks and two model replacements using different network architectures. $AC_{\tau}$ and $AA_{\tau}$ values are computed at task 11, 21. Symbol ``\textendash'' indicates values that cannot be computed.}
    
    \label{tab:iamcl2r_diff_arch}
    \begin{tabular}{l @{\hspace{0.1\tabcolsep}} c @{\hspace{0.3\tabcolsep}} c @{\hspace{1.2\tabcolsep}} c @{\hspace{0.3\tabcolsep}} c @{\hspace{1.2\tabcolsep}} c @{\hspace{0.3\tabcolsep}} c}

    \toprule
         \textsc{method} & $AC_{11}$ & $AA_{11}$ & $AC_{21}$  
         & $AA_{21}$ & $AC$ & $AA$
         \\
    \midrule
        ER baseline & 0.00 &  53.29 &  0.00 & 41.39 & \textendash  & \textendash \\
        FAN~\cite{iscen2020memory} & 0.00 & 52.81 &  0.00 &  40.10  &  \textendash  & \textendash \\
        BCT-ER~\cite{shen2020towards} & 0.00 & 51.31 & 0.00 & 40.00 & \textendash  & \textendash \\
        LCE-ER~\cite{meng2021learning} & 0.00 & 49.93 & 0.00 & 38.90 & \textendash  & \textendash \\
        AdvBCT-ER~\cite{pan2023boundary} & 0.02 & 52.96 & <0.01 & 39.86 & \textendash  & \textendash \\
        CVS~\cite{Wan_2022_CVPR} & 0.00 & 53.30 & 0.00 & 41.43 &  \textendash  & \textendash \\
        $d$-Simplex-FD~\cite{biondi2022cl2r} & 0.16 & 52.47 & 0.36 & 56.88 &  0.34 & 57.58 \\
        $d$-Simplex-HOC (this paper) & \textbf{0.29} &  \textbf{61.56} & \textbf{0.64} & \textbf{67.28} &  \textbf{0.58} & \textbf{71.40}\\
    \bottomrule
    \end{tabular}
\end{table}

\begin{figure}[t]
    \centering
    \begin{subfigure}{\linewidth}
        \centering
        \includegraphics[width=0.85\linewidth]{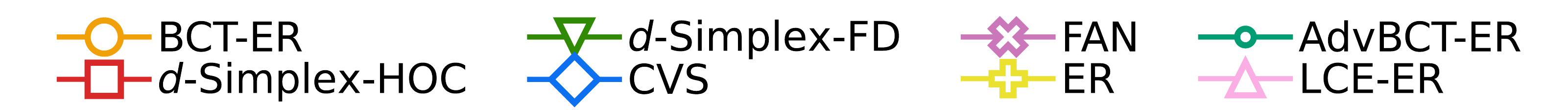}
    \end{subfigure}
    \hspace{-15pt}
    \includegraphics[width=1.01\linewidth]{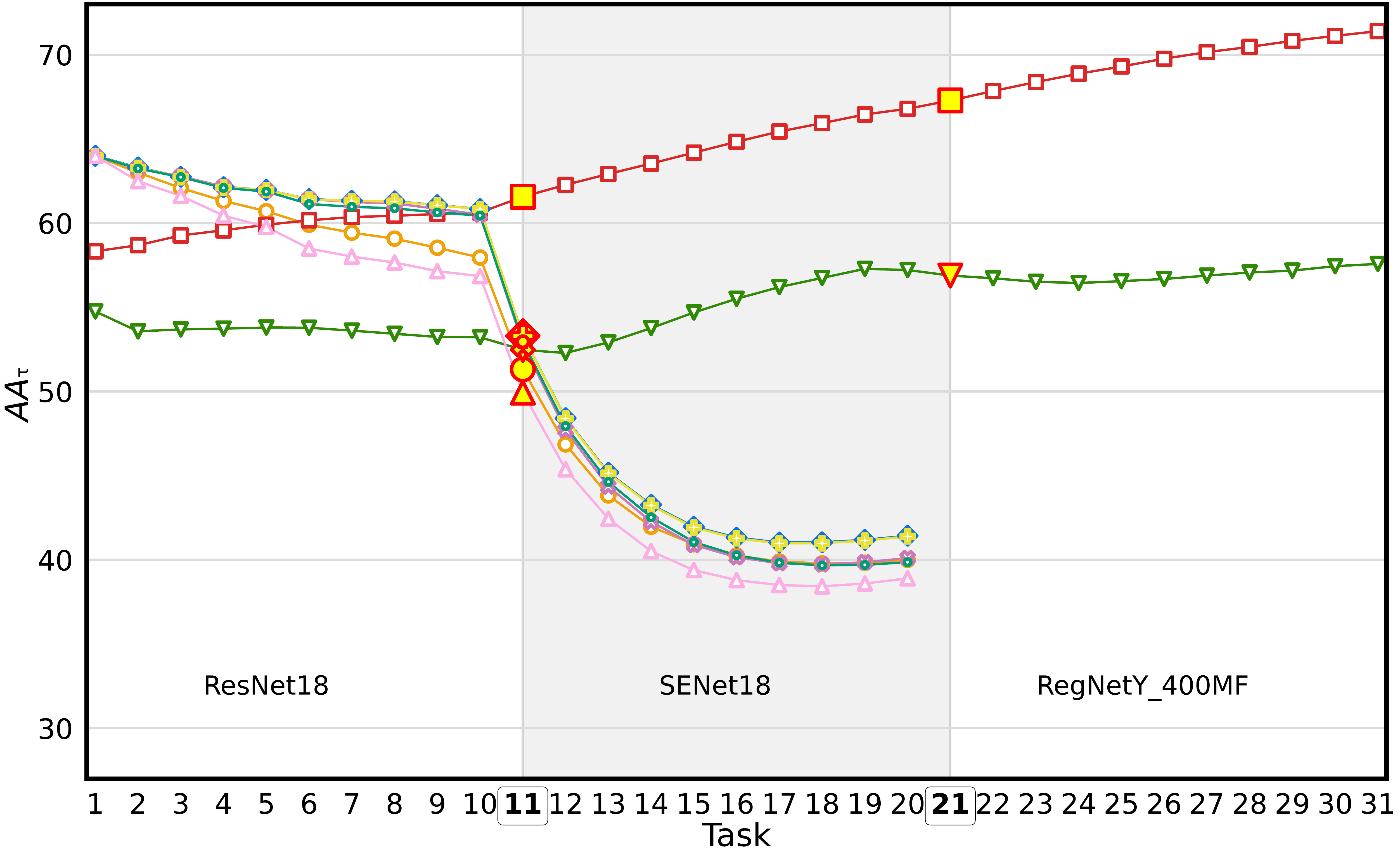}
    \caption{1:N search IAM-CL$^2$R scenario. Average Accuracy up to task $\tau$ ($AA_{\tau}$) 
    with CIFAR100R/10 for 31 tasks and two model replacements using different network architectures. The first tasks after replacements (11 and 21) are marked by boxed task indices.}
    \label{fig:cifar31-iamcl2r-diff-arch}
\end{figure}

\begin{table*}
    \captionof{table}{1:N search CL$^2$R scenario. Ablation of $d$-Simplex-HOC in terms of $AC$, $AA$, $ACA$ with CIFAR100/10 for 16 tasks with varying:
    (a) number of pre-allocated classes $K$;
    (b) $\lambda$ values of Eq.~\ref{eq:total_loss};
    (c) $\rho$ values of Eq.~\ref{eq:CONTRAST2}.
    Values used in the experiments of Sec.~\ref{sec:experiments_cl2r} are marked with the ``(${\scalebox{0.8}{$\spadesuit$}}$)'' symbol. 
    \vspace{-5pt}
    }
    \label{tab:ablation_hyperparams}
\begin{subtable}{.33\linewidth}
    \centering
    \caption{\vspace{-3pt}} \label{tab:pre-alloc_classes}
    \sisetup{detect-weight=true,detect-all,table-format=-1.3,round-mode=places}
    \begin{tabular}{l
                    S[round-precision=2]
                    S[round-precision=2]
                    S[round-precision=2]}
                    
        \arrayrulecolor{black}
        \toprule
        \textsc{$K$} &  $AC$    &  $AA$  & $\mathit{\mathit{ACA}}$ \\
        \midrule
        100 (${\scalebox{0.8}{$\spadesuit$}}$)  &  0.9 & 47.561 & 43.199  \\
        200  &  0.95 & 43.749 & 41.143 \\
        500 &  \textbf{0.98} & 45.354 & 44.06 \\
        1,000 & 0.95 & 46.437 & 43.624 \\
        5,000 &  0.9 & \hspace{-3pt}\textbf{48.69} & 43.675 \\
        10,000 & 0.95 & 48.31 & \hspace{-3pt}\textbf{45.46} \\
        \bottomrule
    \end{tabular}
\end{subtable}
\hspace{2pt}
\begin{subtable}{.33\linewidth}
    \caption{\vspace{-3pt}} \label{tab:lambda}
    \centering    
    \sisetup{detect-weight=true,detect-all,table-format=-1.3,round-mode=places}
    \begin{tabular}{l
                    S[round-precision=2]
                    S[round-precision=2]
                    S[round-precision=2]}
                    
        \arrayrulecolor{black}
        \toprule
        \textsc{$\lambda$} &   $AC$    &  $AA$  & $\mathit{\mathit{ACA}}$  \\
        \midrule
         0 ($\mathcal{L}_{i\textsc{nce}}$)  & \textbf{0.96} & 43 & 40.706 \\
        0.05      &  \textbf{0.96} & 43.812 & 41.62 \\
        0.10 (${\scalebox{0.8}{$\spadesuit$}}$)      & 0.90 & 47.561 & \hspace{-3pt}\textbf{43.20} \\
        0.25      & 0.275 & 48.63 & 13.616 \\
        0.50 & 0.3417 & \hspace{-3pt}\textbf{48.65} & 15.825 \\
        1 ($\mathcal{L}_{\textsc{sce}}$) &  0 & 41.053 & 0 \\
        \bottomrule
    \end{tabular}
\end{subtable}%
\hspace{2pt}
\begin{subtable}{.33\linewidth}
    \caption{\vspace{-3pt}} \label{tab:tau}
    \centering    
    \sisetup{detect-weight=true,detect-all,table-format=-1.3,round-mode=places}
    \begin{tabular}{l
                    S[round-precision=2]
                    S[round-precision=2]
                    S[round-precision=2]}
                    
        \arrayrulecolor{black}
        \toprule
        \textsc{$\rho$} &  $AC$    &  $AA$  & $\mathit{\mathit{ACA}}$ \\
        \midrule
        0.1  & 0.1917 & 39.384 & 8.367 \\
        0.5  & 0.275 & 42.627 & 11.372 \\
        1  & 0.35 & 45.181 & 15.351 \\
        5 (${\scalebox{0.8}{$\spadesuit$}}$)  & \textbf{0.90} & \hspace{-3pt}\textbf{47.56} & \hspace{-3pt}\textbf{43.20} \\
        10 &  0.8333 & 47.324 & 40.98 \\
        25 &  0.5333 & 46.701 & 26.195 \\
        \bottomrule
    \end{tabular}
\end{subtable}
\end{table*}

\noindent\textbf{Model Replacement with Different Network Architectures. \;}
In this case, the improved models are obtained by replacing the original ResNet18 model with SENet18 and then with the RegNetY\_400MF model. 
These are trained from scratch using 300 and 600 ImageNet32 classes, respectively.
he settings are the same as those for CIFAR10/100 in Tab.~\ref{tab:experimental_settings}. The size of the feature vector is 1023 for $d$-Simplex-HOC and $d$-Simplex-FD and equal to the size of the penultimate layer of the networks for the other compared methods, namely 512 for ResNet18 and SENet18, and 384 for RegNetY\_400MF.

Tab.~\ref{tab:iamcl2r_diff_arch} shows values of average compatibility and accuracy for 31 tasks. 
Metrics are computed at the tasks where the pre-trained ResNet18 was replaced by new architectures (11 and 21) and at the final task (31).
Notably, only $d$-Simplex-FD and $d$-Simplex-HOC demonstrate consistent compatibility through the updates with good accuracy.
At task 21, the SENet18 model was replaced by the RegNetY\_400MF model. 
Since the feature vectors before and after the replacement have different sizes, it is impossible for all the methods except $d$-Simplex-HOC and $d$-Simplex-FD to match query vectors obtained with the new representations against gallery features extracted with old representations 
Consequently, starting from task 21, compatibility metrics (\mbox{$AC_{\tau}$}, \mbox{$AA_{\tau}$}) cannot be calculated. 
In Fig.~\ref{fig:cifar31-iamcl2r-diff-arch}, we display plots of the \mbox{$AA_{\tau}$} for in this case. 
For both $d$-Simplex-HOC and $d$-Simplex-FD, the $d$-Simplex acts as a common interface between the RegNetY\_400MF representation and the other representations, allowing to leverage the improvements of the new network architecture.

\section{Ablation Studies}\label{sec:ablation}

In the following, we present ablation studies for $d$-Simplex-HOC in the CL$^2$R and \mbox{IAM-CL$^2$R} scenarios.

\noindent\textbf{Number of Pre-allocated Classes. \;}  
$d$-Simplex fixed classifiers allow to reserve regions in the representation space to pre-allocate future classes. 

We analyze the impact of class pre-allocation on the $d$-Simplex-HOC performance over 16 tasks with CIFAR100/10 in the CL$^2$R scenario. 
Tab.~\ref{tab:pre-alloc_classes} presents the values of $AC$, $AA$, and $\mathit{ACA}$ for 100 classes of CIFAR100, out of 100, 200, 500, 1,000, 5,000 and 10,000 pre-allocated classes.
The results suggest that compatibility and accuracy of $d$-Simplex-HOC are not significantly influenced by the number of pre-allocated classes.
\\

\noindent\textbf{Hyper-parameters Values. \;}
Training of $d$-Simplex-HOC depends on $\lambda$ and $\rho$ hyper-parameters which are used in Eq.~\ref{eq:total_loss} and Eq.~\ref{eq:CONTRAST2}, respectively.

We analyze the impact of these hyper-parameters on the $d$-Simplex-HOC performance over 16 tasks with CIFAR100/10 in the CL$^2$R scenario. 
Tab.~\ref{tab:lambda} and Tab.~\ref{tab:tau} show $AC$, $AA$, and $\mathit{ACA}$ values for different values of $\lambda$ and $\rho$, respectively. From Tab.~\ref{tab:lambda}, $\lambda = 0.1$ achieves the highest performance in terms of $\mathit{ACA}$, so providing the best trade-off between compatibility and accuracy.
From Tab.~\ref{tab:tau}, the best overall performance is obtained with $\rho = 5$. 
However, this may vary across different datasets, as also reported by~\cite{chen2020simple}. We use ($\lambda = 0.1$, $\rho = 5$) for CIFAR100/10, ($\lambda = 0.1$, $\rho = 10$) for TinyImageNet200/20, and ($\lambda = 0.5$, $\rho = 10$) for CUB180/20 in the experiments for the CL$^2$R scenario in Sec. \ref{sec:experiments_cl2r}. 
In the experiments for the IAM-CL$^2$R scenario in Sec. \ref{sec:experiments_iamcl2r}, we use ($\lambda = 0.1$, $\rho = 10$) for CIFAR100R/10 and CelebA.
\\

\begin{figure*}
    \begin{subfigure}{0.32\linewidth}
        \centering
        \includegraphics[width=1\linewidth]{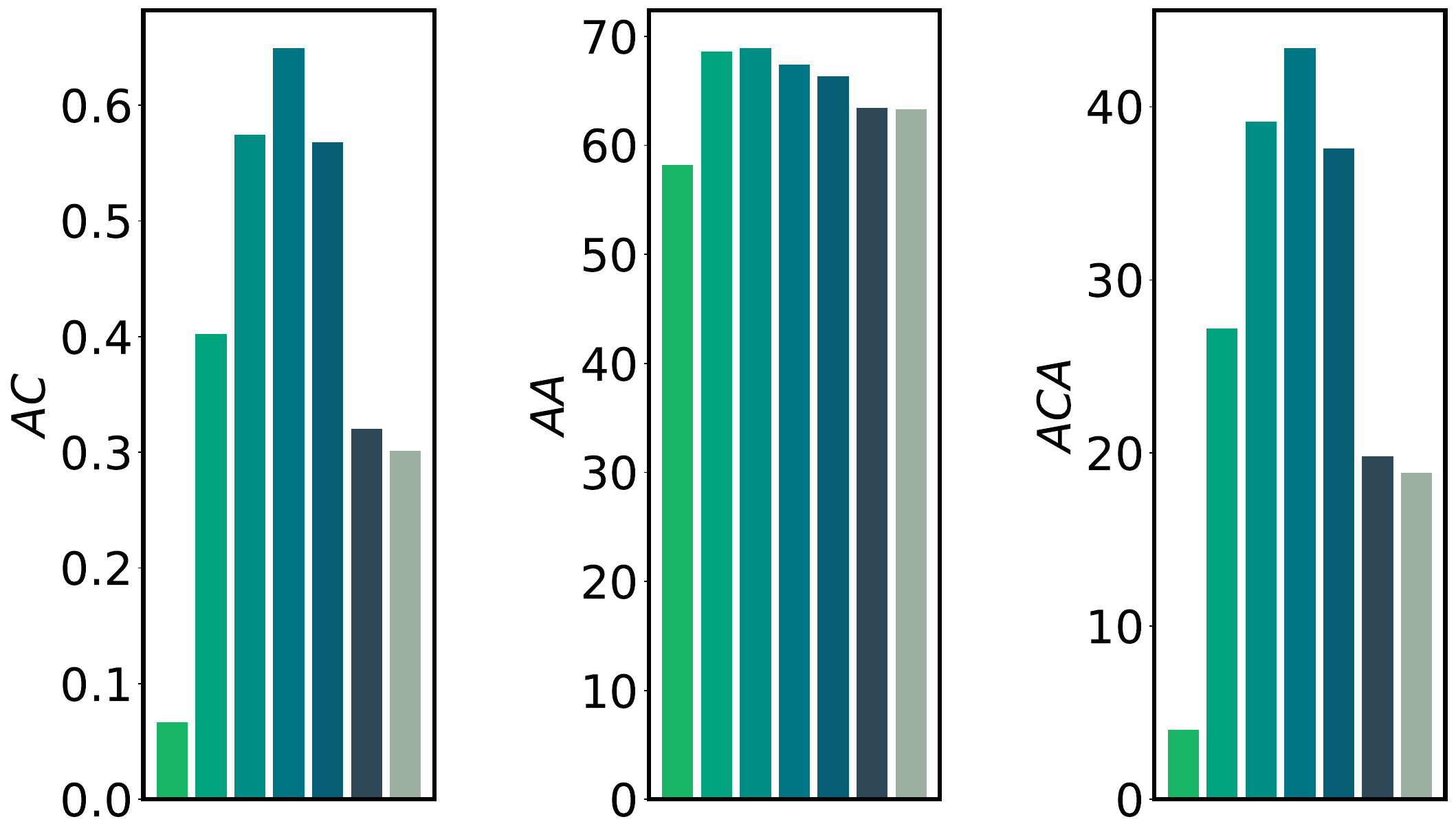}
        \includegraphics[width=1.\linewidth]{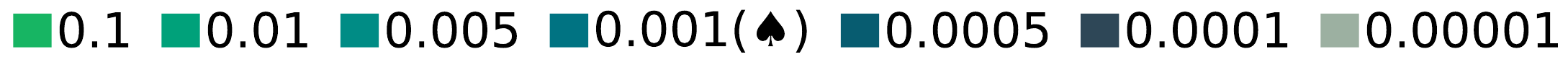}
        \caption{}\label{fig:abl-lr}
    \end{subfigure}
    \hspace{4pt}
    \begin{subfigure}{0.31\linewidth}
        \centering
        \includegraphics[width=\linewidth]{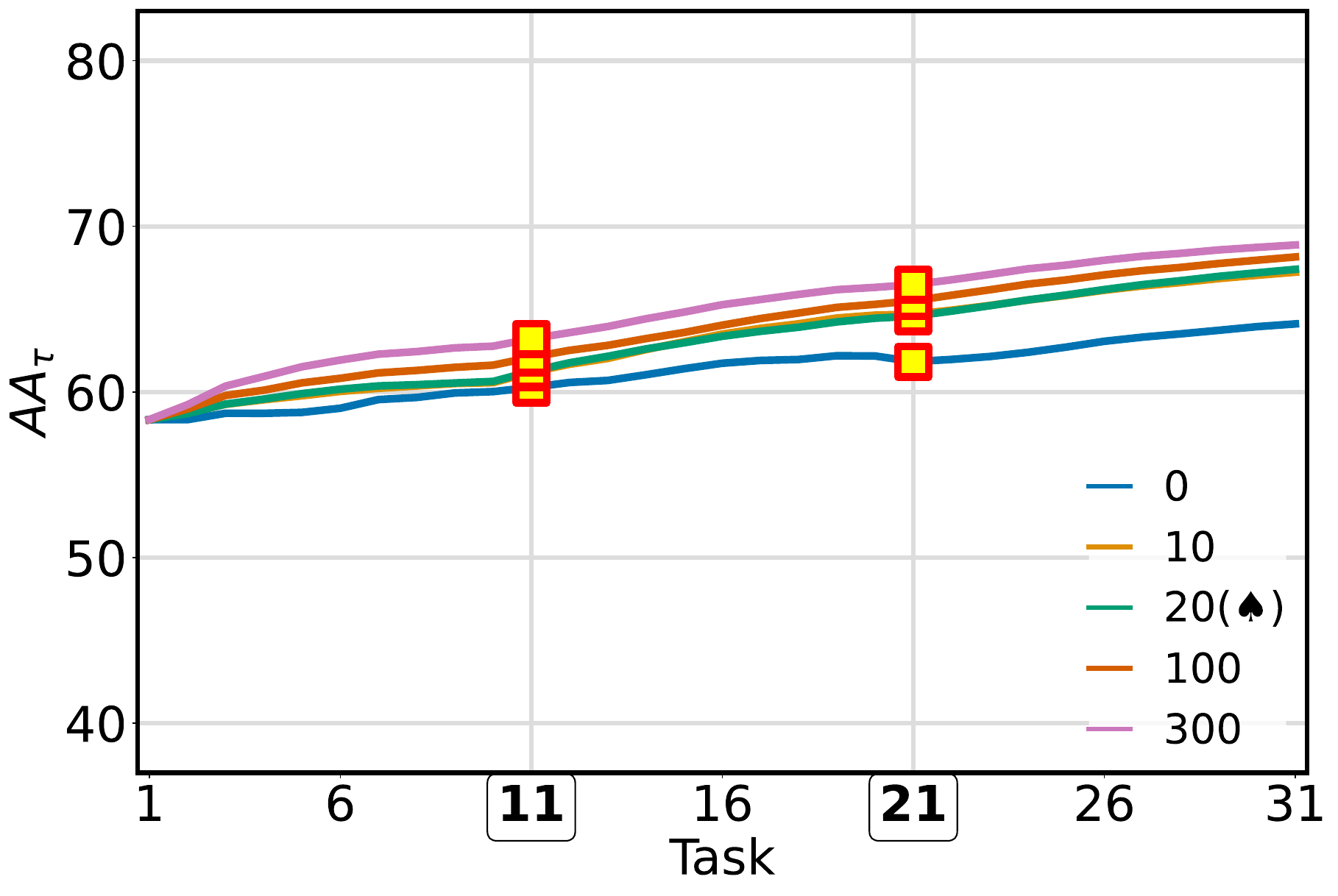}
        \caption{
        }\label{fig:abl-memory-size}
    \end{subfigure}
    \hspace{4pt}
    \begin{subfigure}{0.32\linewidth}
        \centering
        \includegraphics[width=1\linewidth]{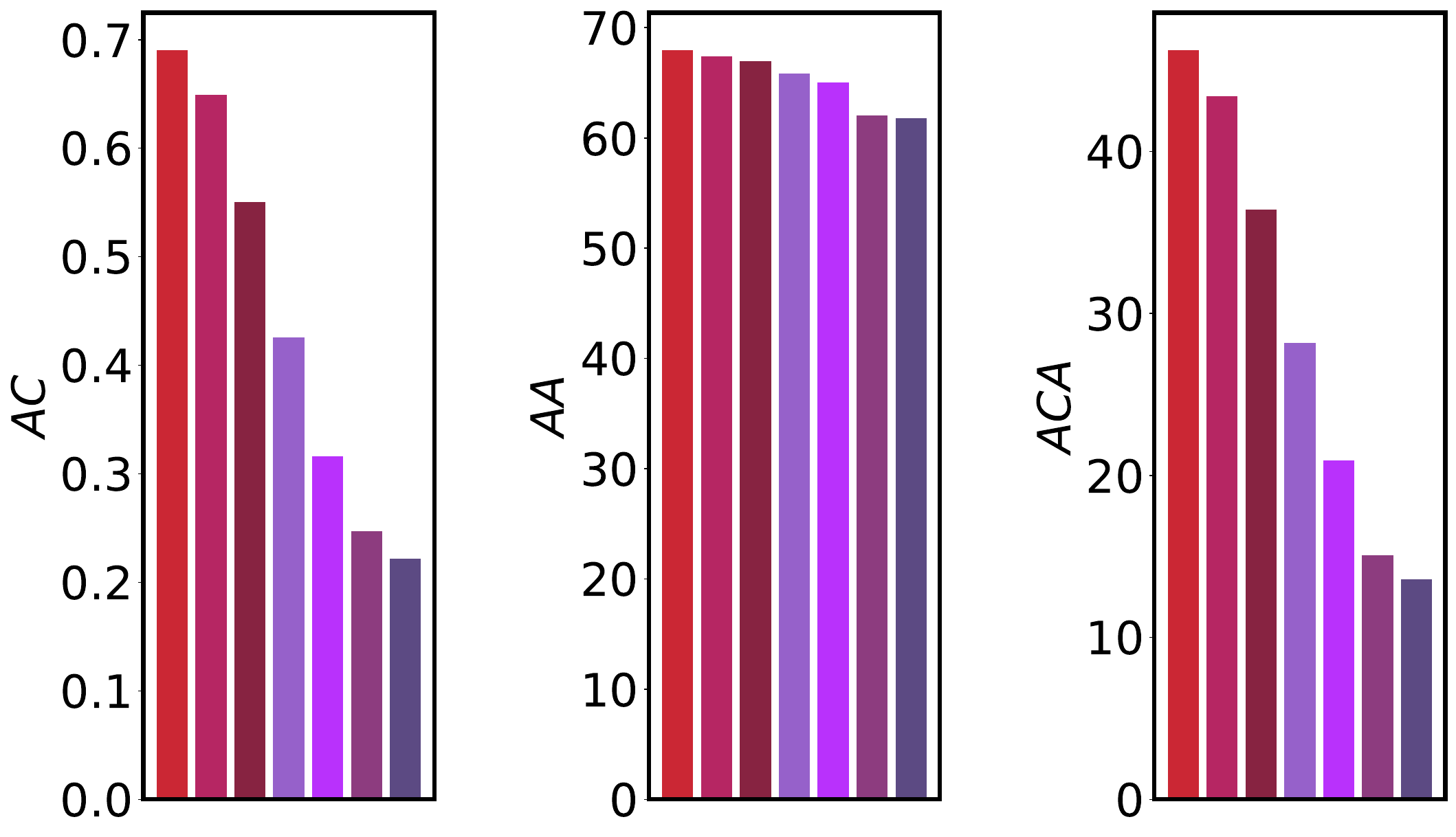}
        \includegraphics[width=0.7\linewidth]{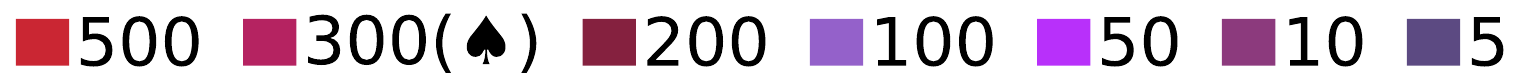}
        \caption{}\label{fig:abl-img-per-class}
    \end{subfigure}
    \caption{1:N search IAM-CL$^2$R scenario. Ablation of $d$-Simplex-HOC with CIFAR100R/10 for 31 tasks and two replacements with models obtained by retraining ResNet18 from scratch with 300 and 600 ImageNet32 classes, respectively.
    (a) $AC$, $AA$, and $ACA$ for different learning rates during fine-tuning;
    (b) $AA_\tau$ for varying numbers of images per class in the experience replay buffer (0 corresponds to the \textit{rehearsal-free} case). The first tasks after replacements (11 and 21) are marked by boxed task indices;
    (c) $AC$, $AA$, and $ACA$ for varying numbers of images per class used in fine-tuning.
    Values used in the experiments of Sec.~\ref{sec:experiments_iamcl2r} are marked with the ``{(${\scalebox{0.7}{$\spadesuit$}}$)}'' symbol.
    }
    \label{fig:enter-label}
\end{figure*}

\noindent\textbf{Learning Rate. \;}  
We analyze the effects of the learning rate in the IAM-CL$^2$R scenario with two model replacements. 
The improved models are obtained by retraining ResNet18 from scratch with 300 and 600 ImageNet32 classes.
Fig.~\ref{fig:abl-lr} reports values of $AC$, $AA$, $\mathit{ACA}$ of models as learned with different learning rates on CIFAR100R/10 with 31 tasks.

It is observable that high learning rates while enabling to learn new tasks, results into lower values of $AA$, $AC$ and $ACA$. The main reason is that the model's representation is strongly modified after the updates. Conversely, low learning rates cause only minimal changes in the representation. This yields higher compatibility but at the cost of accuracy, as the knowledge provided by the new tasks is only weakly exploited.
In the experiments of Sec.~\ref{sec:experiments_iamcl2r}, we use a learning rate equal to $0.001$ as a trade-off value.
\\

\noindent\textbf{Number of Images in the Experience Replay Buffer. \;} 
Model fine-tuning is performed using the current task-set and data in the Experience Replay buffer~\cite{icarl}.

Fig.~\ref{fig:abl-memory-size} shows plots of $AA_{\tau}$ for varying numbers of images per class in the experience replay buffer for 31 tasks with CIFAR100R/10 in the IAM-CL$^2$R scenario with two model replacements.
The improved models are obtained by retraining ResNet18 from scratch with 300 and 600 ImageNet32 classes.
The plots describe the rehearsal-free case (no images are retained in the ER buffer), the case in which all the 300 images of each class are stored in the Experience Replay buffer, and cases with intermediate numbers. 
As expected, accuracy increases with more rehearsal data.
Remarkably, $d$-Simplex-HOC is able to exploit the improvements due to model replacement even in the rehearsal-free case.
\\

\noindent\textbf{Number of Images per Class for Fine-tuning. \;} 
Performance can vary sensibly depending on the number of images per class used for fine-tuning. 

Fig.~\ref{fig:abl-img-per-class} presents values of $AC$, $AA$ and $\mathit{ACA}$ for varying numbers of images per class used for fine-tuning on CIFAR100R/10 with 31 tasks in the IAM-CL$^2$R scenario with two model replacements.
The improved models are obtained by retraining ResNet18 from scratch with 300 and 600 ImageNet32 classes. 

Values are shown for 500, 300, 200, 100, 50, 10, and 5 images per CIFAR100R class. 
We can observe that, while $AC$ and $ACA$ decreases with less images per class, $AA$ maintains relatively high values. 
This is presumably due to the fact that cross-test values, although not high enough to satisfy the compatibility criterion of Eq.~\ref{eq:multistepecc}, are nevertheless still adequate to achieve high $AA$.
In the experiments of Sec. \ref{sec:experiments_iamcl2r}, we use CIFAR100R with 300 images per class.

\section{Conclusions} \label{sec:conclusion_future_directions}

In this paper, we advanced research on compatible representation learning by showing that stationary representations as learned by $d$-Simplex fixed classifiers satisfy the compatibility inequalities of \cite{shen2020towards} in expectation.
Under sequential fine-tuning, we proposed training the model using a $d$-Simplex fixed classifier with a convex combination of the cross-entropy and a contrastive loss, which captures higher-order dependencies in the representation between model updates. We proved that this is equivalent to training with cross-entropy loss using the compatibility inequality as constraints. 

We validate our method with extensive experiments in sequential fine-tuning for visual search, including a new scenario in which a model is occasionally replaced with improved models. 
In this scenario, we showed that, in contrast to all other methods, stationary representations provided by $d$-Simplex fixed classifiers enable uninterrupted retrieval while leveraging the performance improvements of the replacements. 
Our method achieves state-of-the-art performance in all the experiments, particularly for multiple model updates.
\ifCLASSOPTIONcompsoc
  \section*{Acknowledgments}
\else
  \section*{Acknowledgment}
\fi

This work was partially supported by the European Commission under European Horizon 2020 Programme, grant number 951911 - AI4Media.

We acknowledge the CINECA award under the ISCRA initiative, for the availability of high-performance computing resources and support (ISCRA-C ID:~HP10C4TIIM).

\bibliographystyle{unsrt} 
\bibliography{bib}

\makeatletter
\newcommand{\vs}{\bBigg@{1.4}}
\makeatother

\newpage

\vspace{-1cm}
\begin{IEEEbiography}[{\includegraphics[width=1in,height=1.25in,clip,keepaspectratio]{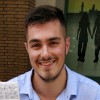}}]{Niccolò Biondi}
Niccolò Biondi is an Assistant Professor with the Department of information engineering and computer science, University of Trento, Italy. He received the M.S. degree (cum laude) in Computer Engineering and the Ph.D. degree (cum laude) in Information Engineering from the University of Florence, Italy, in 2021 and 2025, respectively. His research interests include machine learning, compatible learning, representation learning, lifelong/continual learning, computer vision, and multimodal learning.
\end{IEEEbiography}
\vskip 0pt plus -1fil
\vspace{-1cm}
\begin{IEEEbiography}
[{\includegraphics[width=1in,height=1.25in,clip,keepaspectratio]{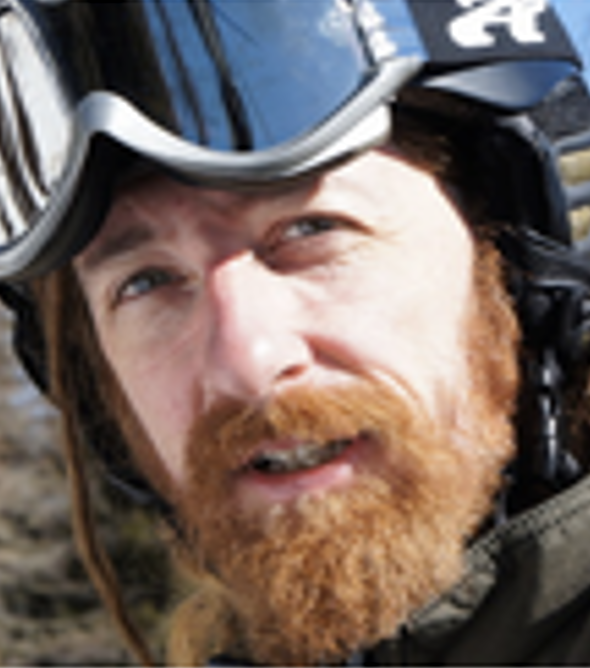}}]{Federico Pernici}
Federico Pernici is Associate Professor at the University of Firenze, Italy. He received the laurea degree in Information Engineering in 2002, the post-laurea degree in Internet Engineering in 2003 and the Ph.D. in Information and Telecommunication Engineering in 2005 from the University of Firenze, Italy. He has been working at the MICC Media Integration and Communication Center as a research assistant since 2002, and has also served as adjunct professor and assistant professor at the University of Firenze. His scientific interests are computer vision and machine learning with a focus on different aspects of visual tracking, incremental learning and representation learning and alignment. He was Associate Editor of Machine Vision and Applications journal. 
\end{IEEEbiography}
\vskip 0pt plus -1fil
\vspace{-1cm}
\begin{IEEEbiography}
  [{\includegraphics[width=1in,height=1.25in,keepaspectratio]{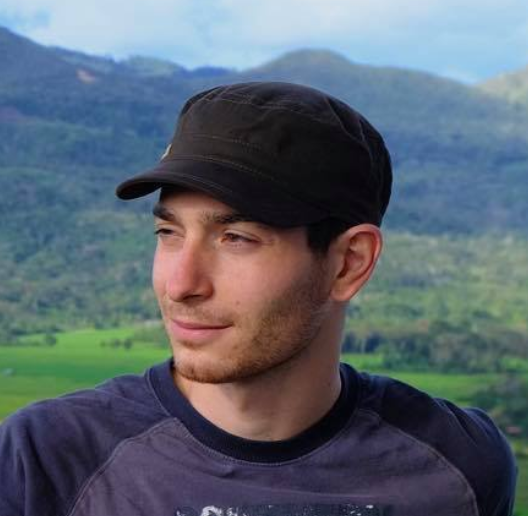}}]{Simone Ricci}
Simone Ricci received the M.S. degree in Computer Engineering from the University of Florence, Italy, in 2018, and the Ph.D. degree in Information Engineering from the University of Florence in 2022. He is currently a Postdoctoral Researcher at the University of Florence. His research interests include pattern recognition and computer vision, with specific focus on incremental learning, compatible representations, compatibility on model prediction, noisy labels, and imbalanced classification.
\end{IEEEbiography}
\vskip 0pt plus -1fil
\vspace{-1cm}
\begin{IEEEbiography}
  [{\includegraphics[width=1in,height=1.25in,clip,keepaspectratio]{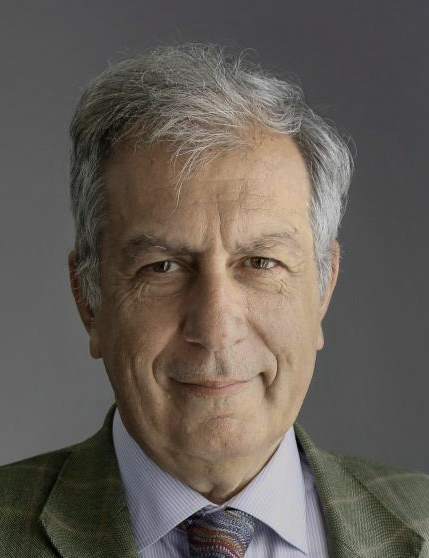}}]{Alberto Del Bimbo}
Prof. Del Bimbo is Emeritus Professor at the University of Firenze, Italy. He is the author of over 350 scientific publications in computer vision and multimedia and principal investigator of technology transfer projects with industry and governments. He was the Program Chair of ICPR 2012, ICPR 2016 and ACM Multimedia 2008, and the General Chair of IEEE ICMCS 1999, ACM Multimedia 2010, ICMR 2011, ECCV 2012 and ICPR2020. He is the General Chair of ACM Multimedia 2021. He is the Editor in Chief of ACM TOMM Transactions on Multimedia Computing Communications and Applications and Associate Editor of Multimedia Tools and Applications and Pattern Analysis and Applications journals. He was Associate Editor of IEEE Transactions on Pattern Analysis and Machine Intelligence, IEEE Transactions on Multimedia and Pattern Recognition and also served as the Guest Editor of many Special Issues in highly ranked journals. Prof. Del Bimbo is the Chair of ACM SIGMM the Special Interest Group on Multimedia. He is IAPR Fellow and ACM Distinguished Scientist and is the recipient of the 2016 ACM SIGMM Award for \emph{Outstanding Technical Contributions to Multimedia Computing Communications and Applications}.
\end{IEEEbiography}

\clearpage

\appendices
\section{Stationarity-Compatibility Theorem} \label{sec:proof-app}

In the following, we demonstrate that stationary representations as learned through the $d$-Simplex fixed classifier imply compatibility as defined in~\cite{shen2020towards}. 
Before proceeding to the proof of Theorem~\ref{theo:compatibility}, two propositions are established.
The first (Proposition~\ref{lemma}) concerns the probability density function of a random point on a hyperspherical cap of the unit hypersphere, while the second (Proposition~\ref{prop3}) addresses the inherent limitations of trainable classifiers to achieve compatibility. 

\begin{customprop}{2}
\label{lemma}
Let $\mathbf{w}_i \in \mathbb{R}^d$ for $i=1, \ldots, n$ be i.i.d. vectors from the uniform distribution on the unit hypersphere. Then the density $f_{\Theta}$ of the polar angle around $\mathbf{w}_i$, evaluated at the expected nearest-neighbor angle $\bar{\theta}(n,d)$ with respect to the other vectors, is given by
\begin{equation}
f_{\Theta}\big(\bar{\theta}(n,d)\big)
=  \frac{\Gamma(d/2)}{\sqrt{\pi} \cdot \Gamma((d-1)/2)} \; \sin ^{d-2} \bar{\theta} (n, d) \label{eq:p_kd}
\end{equation}
where $\Gamma (\cdot)$ is the gamma function.
\end{customprop}

\begin{proof}
The probability $P$ of a random point on a hyperspherical cap centered on $\mathbf{w}_i \in \mathbb{R}^d$ is given by the ratio of the cap surface to the hypersphere's surface area. Approximation of $P$ is obtained as 
$$P=\frac{A_{\text{disc}}}{A}$$ 
where $A$ is the surface area of the hypersphere
\[
A=2 \pi^{d / 2} \frac{R^{d-1}}{\Gamma(d / 2)}
\]
and $A_{\text{disc}}$ is the area of the disc locally approximating the cap around the vector $\mathbf{w}_i$
\[
A_{\text{disc}}=\pi^{(d-1) / 2} \frac{r^{d-1}}{\Gamma((d+1) / 2)}
\]
where $R$ is the radius of the hypersphere and $r$ is the radius of the surface disc area\footnote{Both expression come directly from the $n$-ball volume \mbox{$V_n(R)=\pi^{n/2}R^n/\Gamma(n/2+1)$}: differentiating in $R$ gives the sphere surface area, while setting $n=d-1$ yields the $(d-1)$-ball volume used for the disc surface area.
} (locally approximating the cap).
By substituting these expressions in the expression of $P$, we obtain
\begin{equation}
P=\frac{\Gamma\!\left(\frac{d}{2}\right)}{\sqrt{\pi}\,(d-1)\,\Gamma\!\left(\frac{d-1}{2}\right)}\left(\frac{r}{R}\right)^{\!d-1}.
\label{eq:probability}
\end{equation}

In spherical coordinates, the relationship between $R, r$, and the polar angle $\theta$ is $r=R\sin(\theta)$.
Substituting $r = R\sin\theta$ in Eq.~\ref{eq:probability} and considering the unit hypersphere $R=1$,
we obtain
$$
P\big(\theta\big)= \frac{\Gamma\!\left(\tfrac{d}{2}\right)}{\sqrt{\pi}\,(d-1)\,\Gamma\!\left(\tfrac{d-1}{2}\right)}\,\sin^{\,d-1}\!\theta.
$$
Differentiating $P(\theta)$ with respect to $\theta$ gives the angle density
\begin{equation}
\label{eq:density_theta}
f_{\Theta}(\theta)=\frac{\Gamma\!\left(\tfrac{d}{2}\right)}{\sqrt{\pi}\,\Gamma\!\left(\tfrac{d-1}{2}\right)}\,\sin^{\,d-2}\!\theta\,\cos\theta.
\end{equation}
For small angles, $\cos\theta \approx 1$, so the scaling is governed by $\sin^{\,d-2}\theta$.

Adopting the formulation from~\cite{brauchart2018random,DBLP:conf/cvpr/DengGXZ19}, the expected polar angle $\bar{\theta}$ can be expressed as
\begin{equation}
\bar{\theta} (n, d)= \frac{1}{n^{\frac{2}{d-1}}}  \Gamma \vs( \frac{d}{d-1} \vs) \cdot \left(\frac{\Gamma\left(\frac{d}{2}\right)}{2 \sqrt{\pi}(d-1) \Gamma\left(\frac{d-1}{2}\right)}\right)^{-\frac{1}{d-1}},
\label{eq:theta_kd}
\end{equation}
thereby making explicit its dependence on both the number of vectors $n$ and the hypersphere dimensionality $d$.

Substituting $\bar{\theta}(n,d)$ from Eq. \ref{eq:theta_kd} for $\theta$ in Eq. \ref{eq:density_theta} yields the expression of Eq.~\ref{eq:p_kd}.

\end{proof}

\begin{figure}[t]
    \centering
    \includegraphics[width=0.95\linewidth]{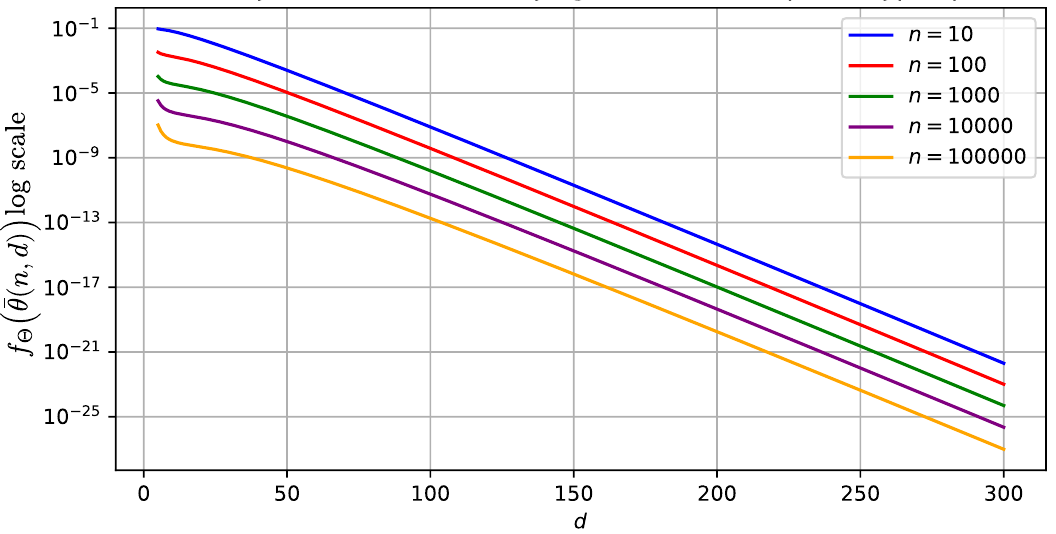}
    \vspace{-5pt}
    \caption{The angle density $f_{\Theta}$ evaluated at the expected nearest-neighbor angle, according to Eq.~\ref{eq:p_kd}. Different curves (logarithmic scale) correspond to varying numbers of vectors ($n$) across a dimension range ($d$).}
    \label{fig:pdisc-label}
\end{figure}

From Proposition~\ref{lemma}, as $d$ and $n$ increase, the expected angle $\bar{\theta}(n,d)$ decreases, and consequently 
$f_{\Theta}(\bar{\theta}(n,d))$ also decreases. Fig.~\ref{fig:pdisc-label} empirically confirms this trend. We leverage this observation in the next proposition.

\begin{customprop}{3} \label{prop3}
The representations learned by trainable classifiers do not satisfy the compatibility conditions when new classes are introduced.
\end{customprop}

\begin{proof}

Following Proposition~\ref{lemma}, re-training the model from scratch using a trainable classifier causes the value of  $f_{\Theta}(\bar{\theta}(n,d))$ to exponentially decrease as both the number of classes and dimensionality increase (see Fig.~\ref{fig:pdisc-label}), thus making unlikely that a class prototype $\mathbf{w}_i$ remains at the previous position. 

Under the assumption of two perfectly aligned models,  Eq.~\ref{eq:theta_kd} implies that adding $m$ new classes yields $\bar{\theta}(n+m,d)<\bar{\theta}(n,d)$, thereby reducing both intra- and inter-class expected distances among features, as also shown by \cite{jiang2024generalize}. 
Consequently, the concentric arrangement of the corresponding class hyperspherical caps across models is no longer preserved, violating the conditions required for optimal compatibility.

In principle, class preallocation could reserve angular space for future classes by assigning a nonzero separation $\alpha_{\text{before}} > 0$ to classes not yet observed (and thus receiving no supervision during training).
However, this setting is equivalent to extreme class imbalance; accordingly, by Theorem 5 in \cite{fang2021exploring}, the prototypes of unsupervised classes collapse during training:
$$
\arccos\langle \mathbf{w}_i,\mathbf{w}_k\rangle \to 0 \;\; \forall\, \mathbf{w}_i\neq \mathbf{w}_k\in U,
$$
where $U$ denotes the set of prototypes with no supervision.
Consequently, the separation after training $\alpha_{\text{after}} :=\min_{U} \arccos\langle \mathbf{w}_i, \mathbf{w}_k\rangle$ tends to zero, and the intended pre-allocated margin is not preserved.
Therefore, even with class pre-allocation, adding $m$ new classes still induces a reduction in both intra- and inter-class expected distances among features.

This shows the inherent limit of trainable classifiers in preventing angle reductions while introducing novel classes, thereby compromising compatibility.

\end{proof}

\begin{customthm}{1}[Stationarity implies Compatibility] 
Let $\mathbf{w}_1, \mathbf{w}_2, \ldots, \mathbf{w}_K$ be the class prototypes of a $d$-Simplex fixed classifier. Consider $\phi_k$ and $\phi_t$ as representation models learned up to the $k$-th and $t$-th tasks, respectively, within this classifier. The numbers of classes learned by each model are denoted by $K_k$ and $K_t$, where $K_k < K_t < K$.
Under the assumption that class hyperspherical caps shrink after model updates, it follows that $\phi_{k}$  and $\phi_{t}$ satisfy on average the compatibility inequalities as in Def.~\ref{def:compatibility-shen}.
\end{customthm}

\begin{figure*}[!ht]
\centering
   \subcaptionbox{Same class \label{fig:distance_sameclass}}{
        \includegraphics[height=4.6cm]{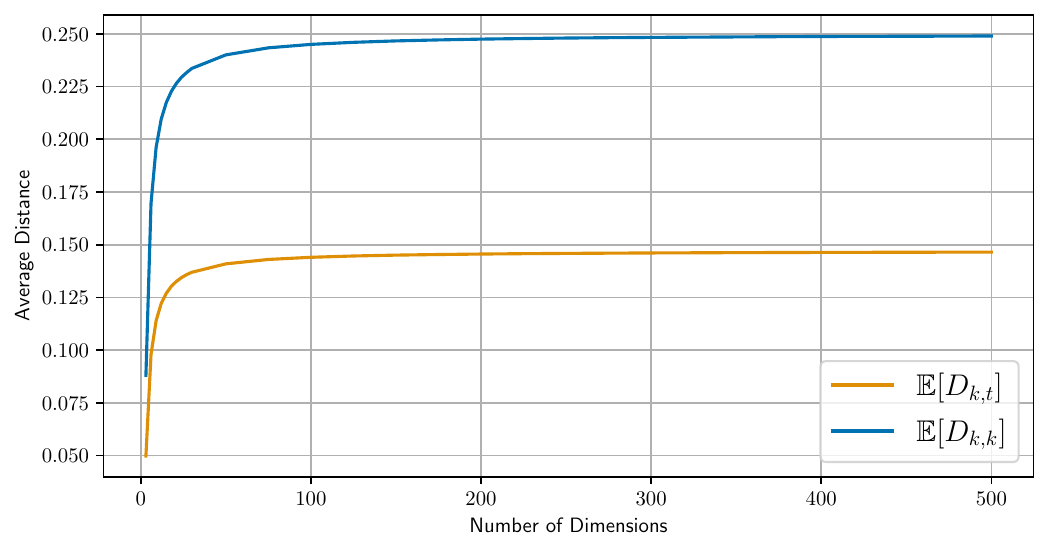}
    }
  \subcaptionbox{Different classes \label{fig:distance_diffclass}}{
        \includegraphics[height=4.6cm]{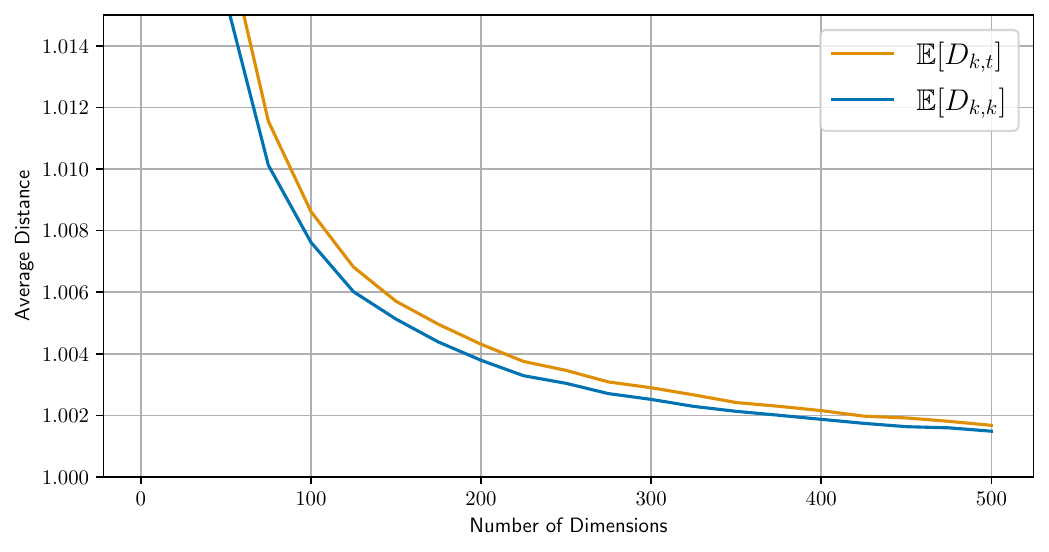}
    }
    \caption{
    Monte Carlo estimates of the expected distances between feature points across different dimensions of the representation space: (a) for the same-class case, $\mathbb{E}[D_{k,t}]$ is lower than $\mathbb{E}[D_{k,k}]$; (b) for the different-class case, $\mathbb{E}[D_{k,t}]$ is higher.     
    }
    \label{fig:distanceHyperSphere}
\end{figure*}

\begin{proof}
Let $\mathbf{H}_k = \phi_k(\mathbf{x})$ and $\mathbf{H}_t = \phi_t(\mathbf{x})$ denote random variables representing the learned representations up to the $k$-th and $t$-th tasks, respectively. Let $f_{\mathbf{H}_k,\mathbf{H}_t}$ be their joint density and $f_{\mathbf{H}_k}$ and $f_{\mathbf{H}_t}$ their marginals. $\mathbf{H}_k$ and $\mathbf{H}_t$ are supported on hyperspherical caps \(\mathcal{C}_k(\mathbf{w}_y, \theta_k^y\)) and \(\mathcal{C}_t(\mathbf{w}_y, \theta_t^y\)) centered on the \(d\)-Simplex prototype \(\mathbf{w}_y\) associated with class \(y\) of angle $\theta^y$ .

From the assumptions it follows: 
(A1) In $d$-Simplex fixed classifiers distances between different class prototypes are the same. Therefore, Eq.~\ref{eq:second} can be verified considering just a pair of classes.
(A2) Being $\mathbf{H}_k$ and $\mathbf{H}_t$ derived from two separate models, they can be regarded as independent from each other. 
(A3) Since the hyperspherical caps shrink after model updates, it results $\theta_t^{y}\le \theta_k^{y}$. \\
According to UFM \cite{mixon2022neural}, we consider the learned representations at each step as independent from each other, so the analysis can concentrate on one class independently of the others.  Cosine distance is used to evaluate distances between hyperspherical representations. 

We define the following random variables representing distances as:
\[  D_{k,t}:=g(\mathbf{H}_k,\mathbf{H}_t),\qquad
D_{k,k}:=g(\mathbf{H}_k,\mathbf{H}_k'),
\]
where $\mathbf{H}_k'$ is an i.i.d. copy of $\mathbf{H}_k$ (i.e., they have the same density $\mathbf{f}_{\mathbf{H}_k}$)  
and $g$ is the cosine distance function.
It follows that:
\[
\mathbb{E}[D_{k,t}]
= \iint\displaylimits_{\mathcal{C}_k^{y_i}\times \mathcal{C}_t^{y_j}}
   g(z_1,z_2)\, f_{\mathbf{H}_k}(z_1)\, f_{\mathbf{H}_t}(z_2)\, dV(z_1)\, dV(z_2),
\]
\[
\mathbb{E}[D_{k,k}]
= \iint\displaylimits_{\mathcal{C}_k^{y_i}\times \mathcal{C}_k^{y_j}}
   g(z_1,z_2)\, f_{\mathbf{H}_k}(z_1)\, f_{\mathbf{H}_k}(z_2)\, dV(z_1)\, dV(z_2).
\]
where $dV(\cdot)$ is the volume element, $\mathcal{C}_k^{y}:=\mathcal{C}_k(\mathbf{w}_y,\theta_k^{y})$, and $\mathcal{C}_t^{y}:=\mathcal{C}_t(\mathbf{w}_y,\theta_t^{y})$. 

Since closed form solutions of these integrals are not available except in the
2D case \cite{fairthorne1964distances}, Monte Carlo integration is employed by sampling over the hyperspherical cap as in \cite{venkatapathi2021n}.
Drawing the following i.i.d. samples: $z_1^{(m)}$ from $f_{\mathbf{H}_k}$ on $\mathcal{C}_k^{y_i}$, $z_2^{(m)} \text{ from } f_{\mathbf{H}_t} \text{ on } \mathcal{C}_t^{y_j}$, $\tilde z_1^{(m)} \text{ from } f_{\mathbf{H}_k} \text{ on } \mathcal{C}_k^{y_j}$, and $\tilde z_2^{(m)} \text{ from } f_{\mathbf{H}_k} \text{ on } \mathcal{C}_k^{y_j}$---with $m = 1, 2, \ldots, M$---Eq.~\ref{eq:first} and Eq.~\ref{eq:second} can be expressed with the empirical estimates as:
\[
\tfrac{1}{M}\sum_{m=1}^M g(z_1^{(m)},z_2^{(m)}) \leq \tfrac{1}{M}\sum_{m=1}^M g(z_1^{(m)},\tilde z_2^{(m)}) \text{ for } y_i =y_j,
\]
\[
\tfrac{1}{M}\sum_{m=1}^M g(\tilde z_1^{(m)},z_2^{(m)}) \leq \tfrac{1}{M}\sum_{m=1}^M g(\tilde z_1^{(m)},\tilde z_2^{(m)}) \text{ for } y_i \neq y_j.
\]

Figs.~\ref{fig:distance_sameclass} and~\ref{fig:distance_diffclass} demonstrate that, under reasonable assumptions, stationary representations learned according to a $d$-Simplex fixed classifier satisfy in expectation compatibility inequalities Eqs.~\ref{eq:first} and~\ref{eq:second} across varying representation dimensions.

\end{proof}

The theorem above extends \cite{biondi2024stationary} by adopting a different geometric assumption. 
In \cite{biondi2024stationary}, representations are modeled as hyperballs and Euclidean distance is used as the similarity metric. 
Compatibility is verified for the same-class case (Eq.~\ref{eq:first}), but not for the different-class case (Eq.~\ref{eq:second}).  
In contrast, we model representations as hyperspherical caps on the unit hypersphere, where similarity is measured via angular separation, namely cosine distance. Unlike Euclidean distance---which is unbounded and grows with inter-class separation---cosine distance is bounded as classes move on the unit hypersphere. 
Under this assumption, when evaluated at the vertices of a $d$-Simplex fixed classifier, the compatibility inequality for the different-class case also holds in expectation (Fig.~\ref{fig:distanceHyperSphere}).

\section{Higher-order Compatibility Loss}\label{sec:appenix_proof_hoc}

\begin{customprop}{1}
Let $\phi_{t}$, $\phi_{t-1}$ be two representation models learned according to a $d$-Simplex fixed classifier where the updated model $\phi_t$ is obtained by fine-tuning $\phi_{t-1}$. Then, training $\phi_t$ with the HOC loss of Eq.~\ref{eq:total_loss} is equivalent to training $\phi_t$ using cross-entropy loss under compatibility constraints.
\end{customprop}

\begin{proof}
In principle, a compatible representation can be learned according to a $d$-Simplex fixed classifier by solving the following constrained optimization problem: 
\begin{equation}\label{eq:constrained_opt_probl_general}
\begin{aligned}
\underset{\phi_t}{\text{argmin}} \quad
&  \mathcal{L}_\textsc{sce}(\phi_t)  
\\
     \text{s.t.} \quad & \dist \big(\phi_k(\mathbf{x}_i), \phi_t(\mathbf{x}_j)\big)   - \dist \big(\phi_k(\mathbf{x}_i),  \phi_k(\mathbf{x}_j)\big) \leq 0 \; 
 \\ & \hspace{60pt} \forall \, (\mathbf{x}_i, \mathbf{x}_j) \text{ such that } y_i = y_j   
\\
& \dist  \big(\phi_k(\mathbf{x}_i),  \phi_t(\mathbf{x}_j)\big) - \dist \big(\phi_k(\mathbf{x}_i), \phi_k(\mathbf{x}_j)\big) \geq 0 \; 
 \\ & \hspace{60pt} \forall \, (\mathbf{x}_i, \mathbf{x}_j) \text{ such that } y_i \neq y_j 
\end{aligned}
\end{equation} 
where the use of the $d$-Simplex cross-entropy loss $\mathcal{L}_\textsc{sce}(\phi_t)$ of Eq.~\ref{eq:loss_ce_simplex} is justified from the inability of trainable classifiers to exploit a pre-allocated representation space (Proposition~\ref{prop3}).

In \cite{pmlr-v98-cotter19a}, the constrained optimization problem of Eq.~\ref{eq:constrained_opt_probl_general} has been shown to be intractable.
However, analysis in \cite{jiang2021churn} revealed that it can be converted into a tractable, unconstrained optimization problem. This is done by using a convex combination of the cross-entropy loss and the Kullback-Leibler divergence function, focusing only on the first of the two constraints. 
In a $d$-Simplex fixed classifier, according to Theorem~\ref{theo:compatibility}, the second constraint is directly satisfied by using the loss $\mathcal{L}_\textsc{sce}(\phi_t)$. 
Therefore, Eq. \ref{eq:constrained_opt_probl_general} can be simplified as:
\begin{equation}\label{eq:constrained_opt_probl}
\begin{aligned}
\underset{\phi_t}{\text{argmin}} \quad
&  \mathcal{L}_\textsc{sce}(\phi_t)  
\\
     \text{s.t.} \quad & \dist \big(\phi_{t-1}(\mathbf{x}_i), \phi_t(\mathbf{x}_j) \big)   - \dist \big(\phi_{t-1}(\mathbf{x}_i),  \phi_{t-1}(\mathbf{x}_j)\big) \leq 0 \; 
 \\ & \hspace{60pt} \forall \, (\mathbf{x}_i, \mathbf{x}_j) \text{ such that } y_i = y_j 
\end{aligned}
\end{equation}
where $\phi_{k}=\phi_{t-1}$. 

The contrastive loss $\mathcal{L}_{i\textsc{nce}} (\phi_t, \phi_{t-1})$ can be approximated, according to~\cite{tian2020contrastive}, as the Kullback-Leibler divergence between the product of the marginals and the joint distribution of $\phi_t$ and $\phi_{t-1}$. 
As a consequence, following \cite{jiang2021churn}, training with the loss $\mathcal{L}_\textsc{hoc}$ of Eq.~\ref{eq:total_loss} is equivalent to the constrained optimization problem of Eq.~\ref{eq:constrained_opt_probl_general}.

\end{proof}

The above equivalence and the ability of the contrastive loss $\mathcal{L}_{i\textsc{nce}} (\phi_t, \phi_{t-1})$ to approximate mutual information between distributions~\cite{tian2020contrastive} enable to capture of higher-order dependencies between
$\phi_t$ and $\phi_{t-1}$ while simultaneously learning compatible representations.

\section{TinyImageNet20 and CUB20 Classes}\label{sec:indices_classes}

In the TinyImageNet-200/20 experiments, we train on TinyImageNet-200 and evaluate on a TinyImageNet-20 test set obtained from the following ImageNet-1k classes:
\begin{equation*}
\resizebox{0.9\hsize}{!}{$%
    \begin{aligned}
    &{\rm n04116512, n03447721, n11939491, n02951585, n02437616,} \\
&{\rm n03538406, n02095889, n02169497, n03127925, n01532829,} \\
&{\rm n03394916, n02727426, n09835506, n02105641, n03598930,} \\
&{\rm n04228054, n03743016, n01582220, n04485082, n03483316} \\
    \end{aligned}
    $}
\end{equation*}

In the CUB-180/20 experiments, we train on CUB-180---containing from the first 180 classes of CUB-200---and evaluate on CUB-20, which comprises the remaining 20 classes of CUB-200, namely:
\begin{table}[!ht]
\setlength{\tabcolsep}{1pt}
\centering
\footnotesize
\begin{tabular}{l l}
181.Worm\_eating\_Warbler,& 182.Yellow\_Warbler, \\ 183.Northern\_Waterthrush,& 184.Louisiana\_Waterthrush, \\ 185.Bohemian\_Waxwing,& 186.Cedar\_Waxwing, \\ 187.American\_Three\_toed\_Woodpecker,& 188.Pileated\_Woodpecker, \\ 189.Red\_bellied\_Woodpecker,& 190.Red\_cockaded\_Woodpecker, \\ 191.Red\_headed\_Woodpecker,& 192.Downy\_Woodpecker, \\ 193.Bewick\_Wren,& 194.Cactus\_Wren, \\     195.Carolina\_Wren,& 196.House\_Wren, \\ 197.Marsh\_Wren,& 198.Rock\_Wren, \\ 199.Winter\_Wren,& 200.Common\_Yellowthroa t. \\
\end{tabular}
\end{table}

\section{\texorpdfstring{$d$}{d}-Simplex fixed classifier PyTorch Code}
\label{sec:simplex:formula}

We provide a GPU-based implementation to generate a $d$-Simplex classifier matrix $\mathbf{W}$ for a given number of pre-allocated classes $K$ that offers faster computation compared to CPU-based implementations~\cite{Pernici_2019_CVPR_Workshops,pernici2021regular,kasarla2022maximum}.

\begin{lstlisting}[language=Python]
def dsimplex_fixed_classifier(K):
    W = torch.zeros((K, K-1))
    W[:-1,:] = torch.eye(K-1)
    W = W.cuda()
    c = torch.sqrt(1 + torch.Tensor([K-1]).cuda())
    W[-1,:] = W[-1,:] + (1 - c) / (K-1)
    W.add_(-torch.mean(W, dim=0))
    W.div_(torch.linalg.norm(W) + 1e-8)
    W.requires_grad = False
    return W
\end{lstlisting}

\end{document}